\documentclass[10pt,twocolumn,letterpaper]{article}

\usepackage[pagenumbers]{cvpr} 

\usepackage{times}
\usepackage{epsfig}
\usepackage{graphicx}
\usepackage{amsmath}
\usepackage{amssymb}

\usepackage{enumitem}
\usepackage{booktabs}

\makeatletter
\DeclareRobustCommand\onedot{\futurelet\@let@token\@onedot}
\def\@onedot{\ifx\@let@token.\else.\null\fi\xspace}

\usepackage{multirow}
\usepackage{algorithm}
\usepackage[noend]{algorithmic}
\usepackage{graphicx, amsmath, amssymb, caption, subcaption, multirow, overpic, textpos}
\usepackage[table]{xcolor}

\def\eg{\emph{e.g}\onedot} 
\def\ie{\emph{i.e}\onedot} 
 
 \def\vs{\emph{vs}\onedot}

\newcommand{\figref}[1]{Fig\onedot~\ref{#1}}
\newcommand{\equref}[1]{Eq\onedot~\eqref{#1}}
\newcommand{\secref}[1]{Sec\onedot~\ref{#1}}
\newcommand{\tabref}[1]{Tab\onedot~\ref{#1}}

\newlength\secmargin
\newlength\paramargin
\newlength\abovetabcapmargin
\newlength\belowtabcapmargin
\newlength\abovefigcapmargin
\newlength\belowfigcapmargin

\usepackage{pifont}

\usepackage{xcolor}

\definecolor{baselinecolor}{gray}{.9}
\newcommand{\baseline}[1]{\cellcolor{baselinecolor}{#1}}

\newlength\savewidth\newcommand\shline{\noalign{\global\savewidth\arrayrulewidth
  \global\arrayrulewidth 1pt}\hline\noalign{\global\arrayrulewidth\savewidth}}
\newcommand{\tablestyle}[2]{\setlength{\tabcolsep}{#1}\renewcommand{\arraystretch}{#2}\centering\footnotesize}

\usepackage[pagebackref,breaklinks,colorlinks]{hyperref}

\usepackage[capitalize]{cleveref}
\crefname{section}{Sec.}{Secs.}
\Crefname{section}{Section}{Sections}
\Crefname{table}{Table}{Tables}
\crefname{table}{Tab.}{Tabs.}

\def\cvprPaperID{****} 
\def\confName{CVPR}
\def\confYear{2022}

\begin{document}

\title{CMT-DeepLab: Clustering Mask Transformers for Panoptic Segmentation}

\author{
Qihang Yu\textsuperscript{1$\ast$}~~~~~~Huiyu Wang\textsuperscript{1}~~~~~Dahun Kim\textsuperscript{2}~~~~Siyuan Qiao\textsuperscript{3}~~~~Maxwell Collins\textsuperscript{3}~~~~~Yukun Zhu\textsuperscript{3}\\
Hartwig Adam\textsuperscript{3}~~~~~~Alan Yuille\textsuperscript{1}~~~~~~Liang-Chieh Chen\textsuperscript{3}\\\textsuperscript{1}Johns Hopkins University~~~~\textsuperscript{2}KAIST~~~~~\textsuperscript{3}Google Research}

\maketitle

\begin{abstract}
We propose Clustering Mask Transformer (CMT-DeepLab), a transformer-based framework for panoptic segmentation designed around clustering.
It rethinks the existing transformer architectures used in segmentation and detection; CMT-DeepLab considers the object queries as cluster centers, which fill the role of grouping the pixels when applied to segmentation.
The clustering is computed with an alternating procedure,
by first assigning pixels to the clusters by their feature affinity,
and then updating the cluster centers and pixel features.
Together, these operations comprise the Clustering Mask Transformer (CMT) layer,
which produces cross-attention that is denser and more consistent with the final segmentation task. CMT-DeepLab improves the performance over prior art significantly by 4.4\% PQ, achieving a new state-of-the-art of 55.7\% PQ on the COCO test-dev set.
\let\thefootnote\relax\footnote{$^\ast$Work done during an internship at Google.}
\end{abstract}

\section{Introduction}
\label{sec:intro}

\begin{figure}[t]
    \centering
    \includegraphics[width=1.0\linewidth]{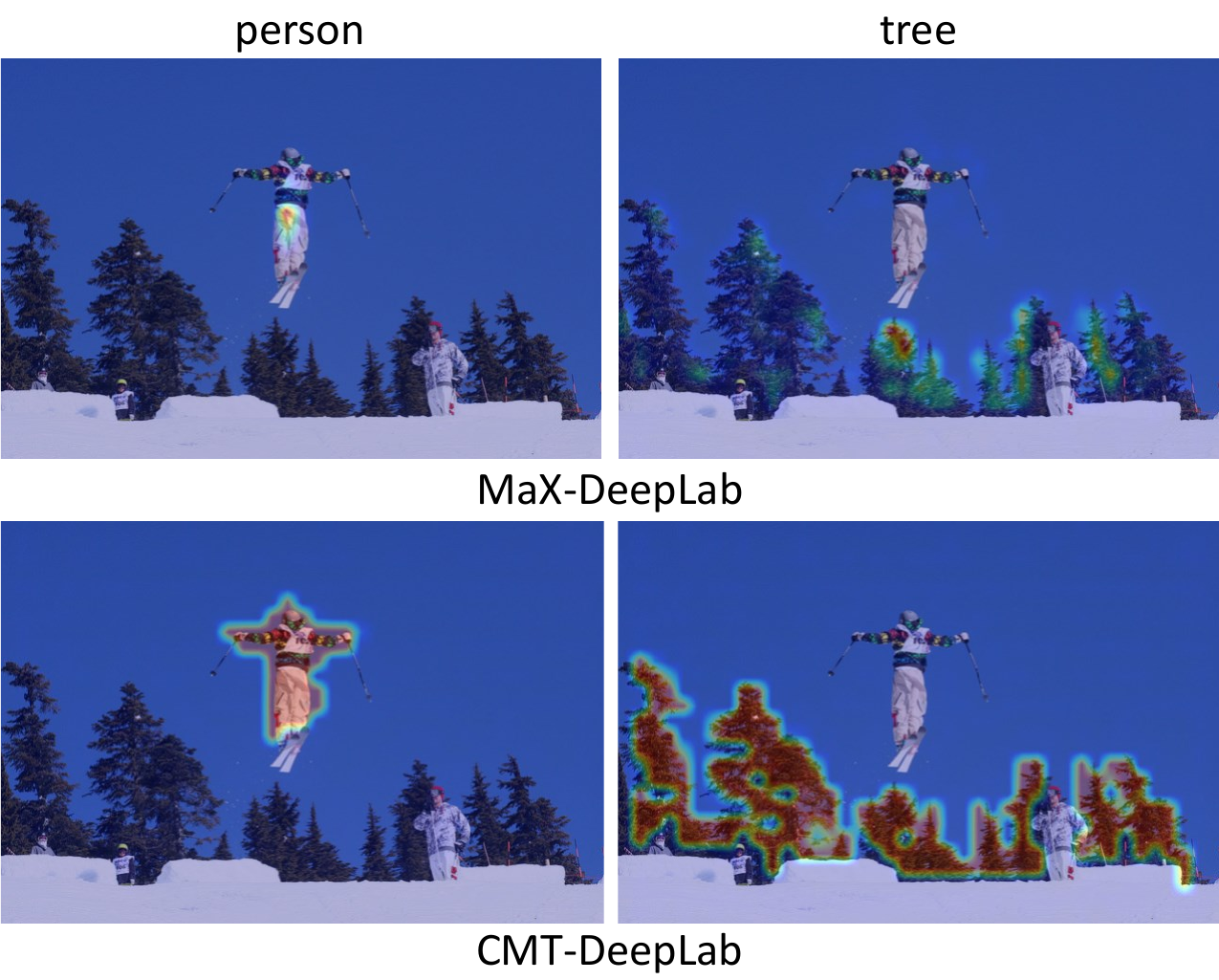}
    \caption{Our CMT-DeepLab generates denser cross-attention maps than MaX-DeepLab~\cite{wang2021max}. The visualization is based on the last transformer layer with averaged multi-head attentions.
    }
    \label{fig:clusterview_of_seg_new}
\end{figure}

Panoptic segmentation~\cite{kirillov2019panoptic}, a recently proposed challenging segmentation task, aims to unify semantic segmentation~\cite{he2004multiscale} and instance segmentation~\cite{hariharan2014simultaneous}. Due to its complicated nature, most panoptic segmentation frameworks~\cite{kirillov2019panoptic,xiong2019upsnet,cheng2019panopticworkshop} decompose the problem into several manageable proxy tasks, such as box detection~\cite{ren2015faster}, box-based segmentation~\cite{he2017mask}, and semantic segmentation~\cite{long2014fully}.

Recently, the paradigm has shifted from the proxy-based approaches to end-to-end systems, since the pioneering work DETR~\cite{carion2020end}, which introduces the first end-to-end object detection method with transformers~\cite{vaswani2017attention}. In their framework, the image features, extracted by a convolutional network~\cite{lecun1998gradient}, are enhanced by transformer encoders. Afterwards, a set of fixed size of positional embeddings, named object queries, interact with the extracted image features through several transformer decoders, consisting of cross-attention and self-attention modules~\cite{bahdanau2014neural}. The object queries, transformed into output embeddings by the decoders, are then {\it directly} used for bounding box predictions.

Along the same direction, end-to-end panoptic segmentation framework~\cite{wang2021max} has been proposed to simplify the panoptic segmentation procedure, avoiding manually designed modules.
The core idea is to exploit a set of object queries conditioned on the inputs to predict a set of pairs, each containing a class prediction and a mask embedding vector. The mask embedding vector, multiplied by the image features, yields a binary mask prediction.
Notably, unlike the box detection task, where the prediction is based on object queries themselves, segmentation mask prediction requires both object queries and pixel features to interact with each other to obtain the results, which consequently incurs different needs when updating the object queries.
To have a deeper understanding towards the role that object queries play, we particularly look into the cross-attention module in the mask transformer decoder, where object queries interact with image features.

Our investigation finds that the update and usage of object queries are performed differently in the transformer-based method for segmentation tasks~\cite{wang2021max}.
Specifically, when updating the object queries, a softmax operation is applied to the image dimension, allowing each query to identify its most similar pixels. On the other hand, when computing the segmentation output, a softmax is performed among the object queries so that each pixel finds its most similar object queries.
The formulation may potentially cause two issues: sparse query updates and infrequent pixel-query communication. First, the object queries are only {\it sparsely} updated due to the softmax being applied to a large image resolution, so it tends to focus on only a few locations (top row in \figref{fig:clusterview_of_seg_new}). Second, the pixels only have {\it one} chance to communicate with the object queries in the final output. The first issue is particularly undesired, since segmentation tasks require dense predictions, and ideally a query should {\it densely} activate all the pixels that belong to the same target. This is different from the box detection task, where object extremities are sufficient (see Fig. 6 of DETR paper~\cite{carion2020end}).

To alleviate the issues, we draw inspiration from the traditional clustering algorithms~\cite{lloyd1982least,achanta2012slic}. In the current end-to-end panoptic segmentation system~\cite{wang2021max}, the final segmentation output is obtained by assigning each pixel to the object queries based on the feature affinity, similar to pixel-cluster assignment step in~\cite{lloyd1982least,achanta2012slic}. The observation motivates us to rethink the transformer-based methods from the clustering perspective by considering the object queries as cluster centers.
We therefore propose to additionally perform the cluster-update step, where the centers are updated by pooling pixel features based on the clustering assignment, when updating the cluster centers (\ie, object queries) in the cross-attention module. As a result, our model generates denser attention maps (bottom row in \figref{fig:clusterview_of_seg_new}). We also utilize the pixel-cluster assignment to update the pixel features within each transformer decoder, enabling frequent communication between pixel features and cluster centers.

Additionally, we notice that in the cross-attention module, pixel features are treated as in ``bag of words"~\cite{lazebnik2006beyond}, while the location information is not well utilized. To resolve the issue, we propose to adopt a dynamic position encoding conditioned on the inputs for {\it location-sensitive} clustering. 
We explicitly predict a reference mask consisting of a few points for each cluster center. The {\it location-sensitive} clustering is then achieved by adding location information to pixel features and cluster
centers via the coordinate convolution~\cite{liu2018intriguing} at the beginning of each transformer decoder.

\begin{figure}[t]
    \centering
    \includegraphics[width=0.9\linewidth]{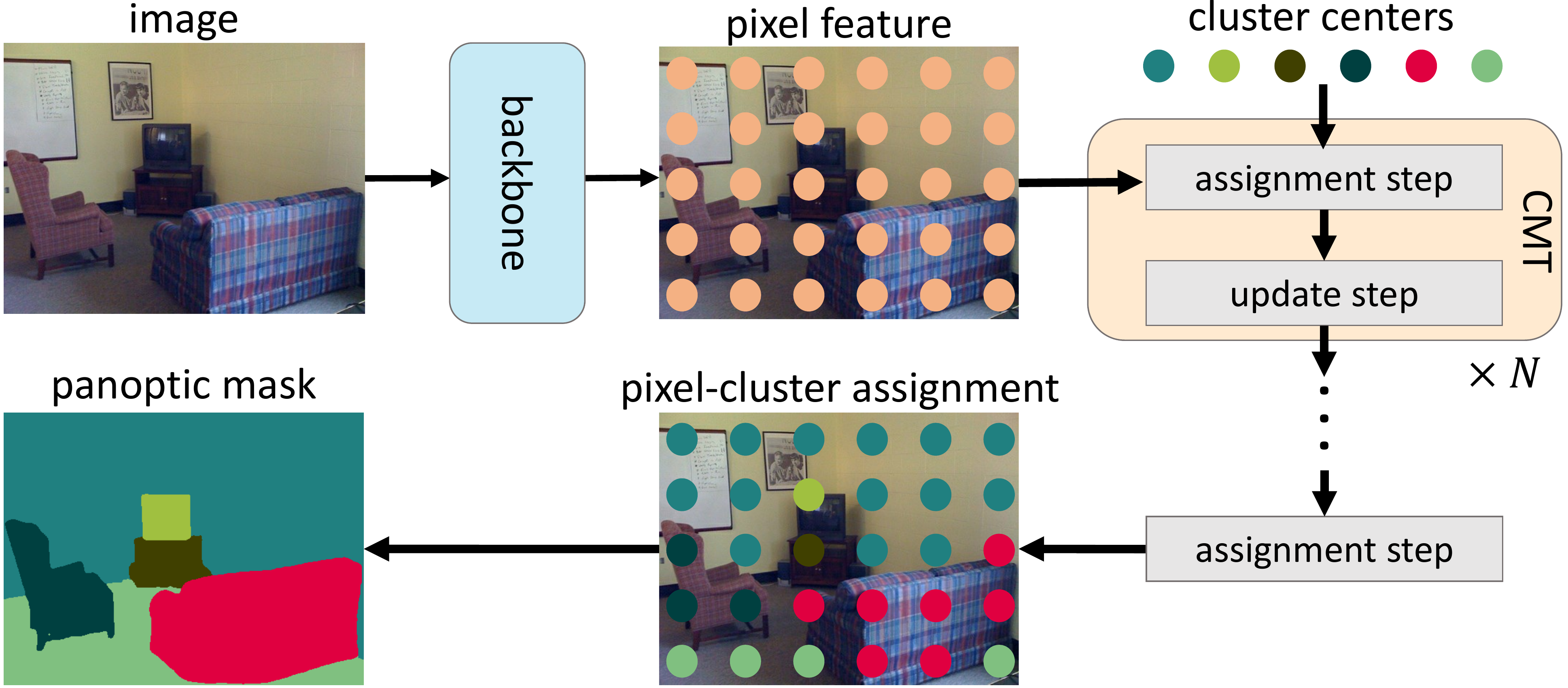}
    \caption{Panoptic segmentation from a clustering perspective. In the proposed Clustering Mask Transformer (CMT) layer, pixels are assigned to cluster centers based on the feature affinity, and the clustering results are used to update both pixel features and cluster centers. After several CMT layers, a refined pixel-cluster assignment is obtained, resulting in the final panoptic mask.
    }
    \label{fig:clusterview_of_seg}
\end{figure}

Combining all the proposed components results in our CMT-DeepLab, which reformulates and further improves the previous end-to-end panoptic segmentation system~\cite{wang2021max} from the traditional clustering perspective. The panoptic segmentation result is naturally obtained by assigning each pixel to its most similar cluster center based on the feature affinity (\figref{fig:clusterview_of_seg}). In the Clustering Mask Transformer (CMT) module, the pixel features, cluster centers, and pixel-cluster assignments are updated in a manner similar to the clustering algorithms~\cite{lloyd1982least,achanta2012slic}. As a result, without bells and whistles, our proposed CMT-DeepLab surpasses its baseline MaX-DeepLab~\cite{wang2021max} by 4.4\% PQ and achieves 55.7\% PQ on COCO panoptic {\it test-dev} set~\cite{lin2014microsoft}.

\section{Related Works}
\label{sec:related}

\noindent\textbf{Transformers.}
Transformer~\cite{vaswani2017attention} variants~\cite{kitaev2020reformer,wang2020linformer,luong2015effective,child2019generating,beltagy2020longformer,zaheer2020big,gupta2020gmat,ainslie2020etc} have advanced the state-of-the-art in many natural language processing tasks~\cite{devlin2019bert,shaw2018self,dai2019transformer} by capturing relations across modalities~\cite{bahdanau2014neural} or in a single context (self-attention)~\cite{cheng2016long,vaswani2017attention}. In computer vision, transformers are either combined with CNNs~\cite{wang2018non,buades2005non} or used as standalone models~\cite{parmar2019stand,hu2019local,wang2020axial,dosovitskiy2020image,liu2021swin}. Both classes of methods have boosted various vision tasks, such as image classification~\cite{chen20182,bello2019attention, parmar2019stand,hu2019local,li2020neural,wang2020axial,dosovitskiy2020image,liu2021swin}, object detection~\cite{wang2018non,shen2021efficient,parmar2019stand,hu2018relation,carion2020end,zhu2020deformable}, semantic segmentation~\cite{chen2016attention,zhao2018psanet,huang2019ccnet,fu2019dual,zhu2019asymmetric,zhu2019empirical}, video recognition~\cite{wang2018non,chen20182,kim2022tubeformer}, image generation~\cite{parmar2018image,ho2019axial}, and panoptic segmentation~\cite{wang2020axial}.

\vspace{1ex}
\noindent\textbf{Proxy-based Panoptic Segmentation.}
Most panoptic segmentation methods rely on proxy tasks, such as object bounding box detection. For example, Panoptic FPN~\cite{kirillov2019panoptic} follows a box-based approach that detects object bounding boxes and predicts a mask for each box, usually with a Mask R-CNN~\cite{he2017mask} and FPN~\cite{lin2017feature}. Then, the instance segments (`thing') and semantic segments (`stuff')~\cite{chen2018deeplabv2} are fused by merging modules~\cite{li2018learning,li2018attention,porzi2019seamless,liu2019e2e,yang2020sognet,xiong2019upsnet,li2020unifying} to generate panoptic segmentation. Other proxy-based methods typically start with semantic segments~\cite{deeplabv12015,chen2017deeplabv3,deeplabv3plus2018} and group `thing' pixels into instance segments with various proxy tasks, such as instance center regression~\cite{kendall2018multi,uhrig2018box2pix,neven2019instance,yang2019deeperlab,cheng2019panoptic,wang2020axial,li2021fully}, Watershed transform \cite{vincent1991watersheds,bai2017deep,bonde2020towards}, Hough-voting \cite{ballard1981generalizing,leibe2004combined,bonde2020towards}, or pixel affinity \cite{keuper2015efficient,liu2018affinity,sofiiuk2019adaptis,gao2019ssap,bonde2020towards}. DetectoRS~\cite{qiao2020detectors} achieved the state-of-the-art in this category with recursive feature pyramid and switchable atrous convolution.
Recently, DETR~\cite{carion2020end} extended the proxy-based methods with its transformer-based end-to-end detector.

\vspace{1ex}
\noindent\textbf{End-to-end Panoptic Segmentation.}
Along the same direction, MaX-DeepLab~\cite{wang2021max} proposed an end-to-end strategy, in which class-labeled object masks are directly predicted and are trained by Hungarian matching the predicted masks with ground truth masks.
In this work, we improve over MaX-DeepLab by approaching the pixel assignment task from a clustering perspective.
Concurrent with our work, Segmenter~\cite{strudel2021segmenter} and MaskFormer~\cite{cheng2021per} formulated an end-to-end strategy from a mask classification perspective, same as MaX-DeepLab~\cite{wang2021max}, but extends from panoptic segmentation to semantic segmentation.

\section{Method}

\begin{figure}[t]
    \centering
    \scalebox{1.0}{
    \includegraphics[width=0.8\linewidth]{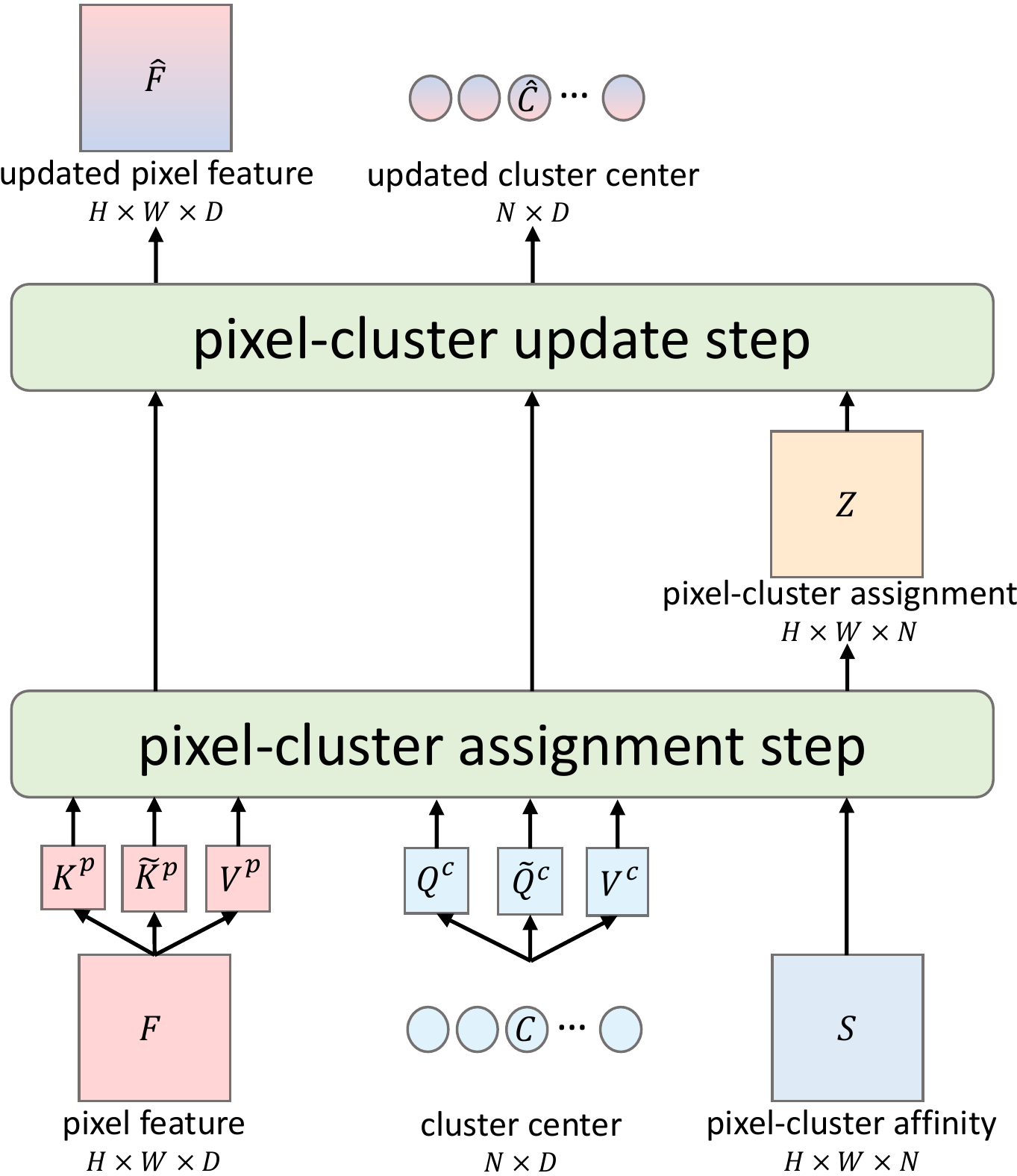}
    }
    \caption{A visual illustration of Clustering Mask Transformer layer, where three variables are updated in a dynamic manner based on the clustering results: pixel features, cluster centers, and pixel-cluster affinity. Details of assignment and update steps are illustrated in \figref{fig:detailed_clusteringattn}.}
    \label{fig:clusteringattn}
\end{figure}

Herein, we firstly introduce recent transformer-based methods~\cite{wang2021max} for end-to-end panoptic segmentation. Our observation reveals a difference between the cross-attention and final segmentation output regarding the way that they utilize object queries. We then propose to resolve it
with a clustering approach,
resulting in our proposed Clustering Mask Transformer (CMT-DeepLab), as shown in \figref{fig:clusteringattn} and \figref{fig:detailed_clusteringattn}. In the following parts, object queries and cluster centers refer to the same learnable embedding vectors and we use them interchangeably for clearer representation.

\subsection{Transformers for Panoptic Segmentation}
\noindent\textbf{Problem Statement.} Panoptic segmentation aims to segment the input image $\mathbf{I} \in \mathbb{R}^{H \times W \times 3}$ into a set of non-overlapping masks as well as the semantic labels for the corresponding masks:
\begin{equation}
\{y_i\}_{i=1}^K = \{(m_i, c_i)\}_{i=1}^K \,.
\end{equation}
The $K$ ground truth masks $m_i \in {\{0,1\}}^{H \times W}$ do not overlap with each other, \ie, $\sum_{i=1}^{K} m_i \leq 1^{H \times W}$, and $c_i$ denotes the ground truth class label of mask $m_i$.

Inspired by DETR~\cite{carion2020end}, several transformer-based end-to-end panoptic segmentation methods~\cite{wang2021max} have been proposed recently, which directly predict $N$ masks and their semantic classes. $N$ is a fixed number and $N\geq K$.
\begin{equation}
\{\hat{y_i}\}_{i=1}^N = \{(\hat{m_i}, \hat{p}_{i}(c))\}_{i=1}^N,
\end{equation}
where $\hat{p}_{i}(c)$ denotes the predicted semantic class confidence for the corresponding mask, including `thing' classes, `stuff' classes, and the void class $\varnothing$.

To predict these $N$ masks, $N$ object queries are utilized to aggregate information from the image features through a transformer decoder, which consists of self-attention and cross-attention modules. The object queries and image features interact with each other in the cross-attention module:
\begin{align}
\label{eq:detr_transformer_update}
\hat{\mathbf{C}}& = \mathbf{C} + \operatornamewithlimits{softmax}_{HW}(\mathbf{Q}^{c} \times (\mathbf{K}^{p})^{\mathrm{T}}) \times \mathbf{V}^{p},
\end{align}
where $\mathbf{C} \in \mathbb{R}^{N \times D}$ refers to object queries with $D$ channels, and $\hat{\mathbf{C}}$ denotes the updated object queries. We use the underscript to represent the axis for softmax, and superscripts $p$ and $c$ to indicate the feature projected from the image features and object queries, respectively. $\mathbf{Q}^c \in \mathbb{R}^{N \times D}, \mathbf{K}^p \in \mathbb{R}^{HW \times D}, \mathbf{V}^p \in \mathbb{R}^{HW \times D} $ stand for the linearly projected features for query, key, and value. For simplicity, we ignore multi-head attention and feed-forward network (FFN) in the equation.

The object queries, updated by multiple transformer decoders, are employed as dynamic convolution weights (with kernel size $1\times 1$)~\cite{jia2016dynamic,tian2020conditional,wang2020solov2} to obtain the prediction $\mathbf{Z} \in \mathbb{R}^{HW \times N}$ that consists of $N$ binary masks. That is,
\begin{align}
\label{eq:detr_final_output}
\mathbf{Z}& = \operatornamewithlimits{softmax}_{N}(\mathbf{F} \times \mathbf{C}^{\mathrm{T}}),
\end{align}
where $\mathbf{F} \in \mathbb{R}^{HW \times D}$ refers to the extracted image features.

\subsection{Current Issues and New Clustering Perspective}
\label{sec:issues_and_clustering}
Even though effective, the transformer-based architectures were originally designed for object detection~\cite{carion2020end} and thus they do not naturally deal with segmentation masks. Specifically, they use different formulations for the object query updates and the segmentation specific output head. To be precise, both the update of object queries (\equref{eq:detr_transformer_update}) and final output (\equref{eq:detr_final_output}) are based on their corresponding feature affinity (\ie, $\mathbf{Q}^{c} \times (\mathbf{K}^{p})^{\mathrm{T}}$ and $\mathbf{F} \times \mathbf{C}^{\mathrm{T}}$). However, the following softmax operations are applied along different dimensions. To update the object queries, the softmax is applied to the image spatial dimension ($HW$) with the goal to identify the most similar pixels for each query. On the other hand, to obtain the final output, the softmax is performed among the object queries ($N$) so that each pixel finds its most similar object queries. The inconsistency potentially causes two issues. First, the object queries are only {\it sparsely} updated due to the softmax operated along a large spatial dimension, tending to focus on only a few locations (\figref{fig:clusterview_of_seg_new}). Second, the output update is only performed {\it once} in the end, and therefore the pixels only have one chance to receive the information passed from the object queries.

To alleviate the issues, we take a closer look at \equref{eq:detr_final_output}, which assigns each pixel to the object queries based on the feature affinity. This is, in fact, very similar to typical clustering methods~\cite{lloyd1982least,achanta2012slic} (particularly, the pixel-cluster assignment step). This observation motivates us to rethink the transformer-based methods from the typical clustering perspective~\cite{zhu1996region,achanta2012slic} by considering the object queries $\mathbf{C}$ as cluster centers. With the clustering perspective in mind, we re-interpret \equref{eq:detr_final_output} as the pixel-cluster assignment. This interpretation naturally inspires us to perform a cluster-update step where the cluster centers are updated by pooling pixel features based on the clustering assignment, \ie, $\mathbf{Z}^{\mathrm{T}} \times \mathbf{F} = (\operatornamewithlimits{softmax}_{N}(\mathbf{F} \times \mathbf{C}^{\mathrm{T}}))^{\mathrm{T}} \times \mathbf{F}$.

We propose to extend the formulation to a transformer decoder module, whose query, key, and value are obtained by linearly projecting the image features and cluster centers:

\begin{align}
\label{eq:cluster_transformer_update}
\hat{\mathbf{C}}& = \mathbf{C} + (\operatornamewithlimits{softmax}_{N}(\mathbf{\tilde{K}}^{p} \times (\mathbf{\tilde{Q}}^{c})^{\mathrm{T}}))^{\mathrm{T}} \times \mathbf{V}^{p}.
\end{align}

Comparing \equref{eq:detr_transformer_update} and \equref{eq:cluster_transformer_update}, we have the query $\mathbf{\tilde{Q}}^{c}$ and key $\mathbf{\tilde{K}}^{p}$ coming from another linear projection, and the softmax is performed along the cluster center dimension.

In the following subsection, we detail how the clustering perspective alleviates the issues of current transformer-based methods. In the discussion, we use object queries and cluster centers interchangeably.

\begin{figure}[tb]
    \centering
    \scalebox{1.0}{
    \includegraphics[width=1.0\linewidth]{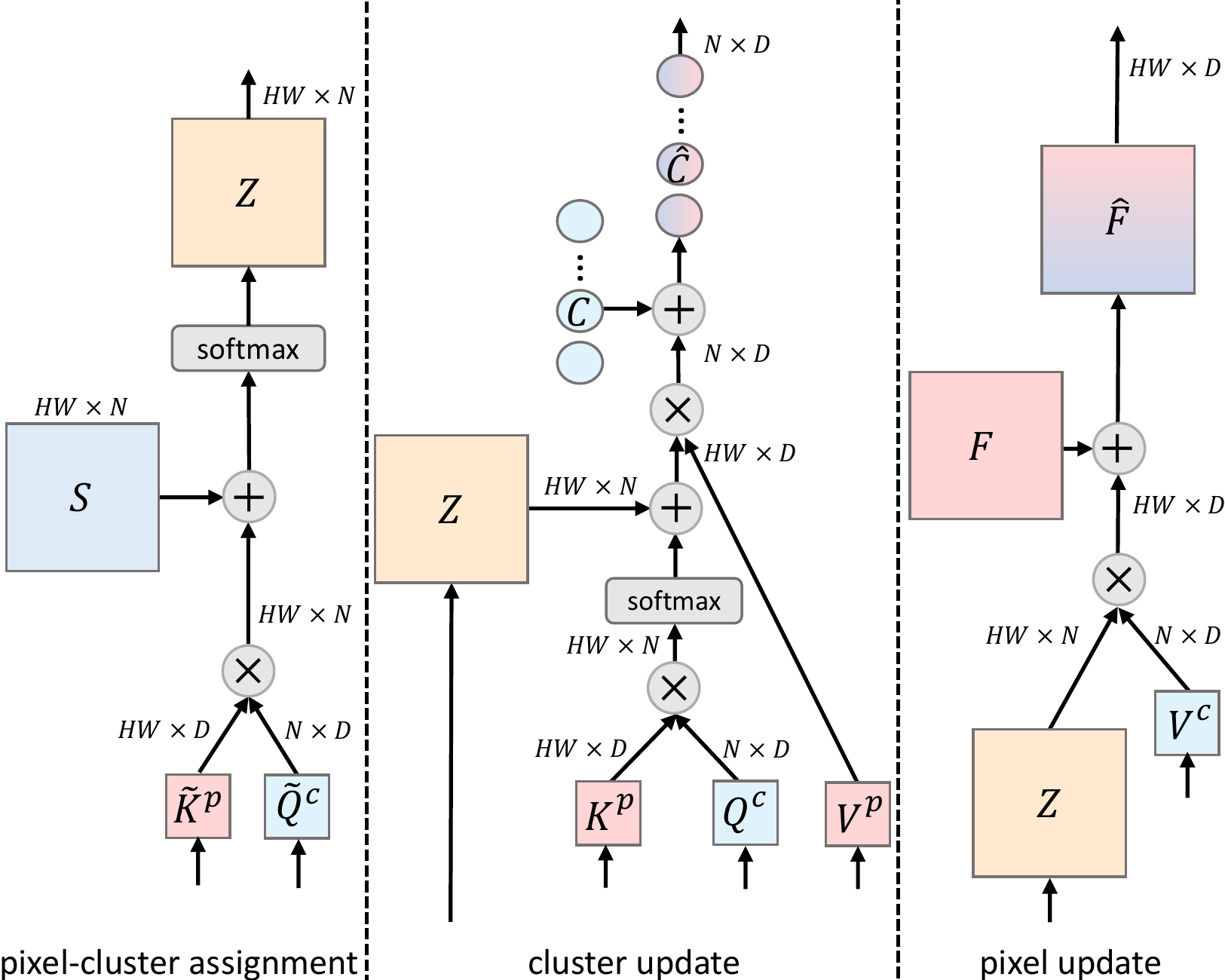}
    }
    \caption{Detailed visual illustration of pixel-cluster assignment (left), cluster centers update (middle), and pixel features update (right). The tensor shapes are specified for illustration.}
    \label{fig:detailed_clusteringattn}
\end{figure}

\subsection{Clustering Mask Transformers}
In this subsection, we redesign the cross-attention in the transformer decoder from the clustering perspective, aiming to resolve the issues raised in \secref{sec:issues_and_clustering}.

\vspace{0.5ex}
\noindent\textbf{Residual Path between Cluster Assignments.} Similar to other designs~\cite{carion2020end}, we stack the transformer decoder multiple times. To facilitate the learning of pixel-cluster assignment, we add a residual connection~\cite{he2016deep} between clustering results including the final segmentation result. That is,

\begin{align}
\label{eq:residual_cluster_transformer_perform_clustering}
\mathbf{Z} &= \operatornamewithlimits{softmax}_{N}(\mathbf{S} + \mathbf{\tilde{K}}^{p} \times (\mathbf{\tilde{Q}}^{c})^{\mathrm{T}}),
\end{align}
where $\mathbf{S} \in \mathbb{R}^{HW \times N}$ is the affinity logits between linearly projected pixel features and cluster centers in the previous decoder (left panel of \figref{fig:detailed_clusteringattn}). We emphasize that since our clustering results have the same format as the segmentation output, we are able to add residual connections between them, which is further supervised by the ground-truths.

\vspace{0.5ex}
\noindent\textbf{Solution to Sparse Query Update.} We propose a simple and effective solution to avoid the sparse query update by combining the proposed clustering center update (\ie, \equref{eq:cluster_transformer_update}) with the original cross-attention (\ie, \equref{eq:detr_transformer_update}), resulting in
\begin{equation}
\begin{split}
\label{eq:cluster_update_both_eq}
\hat{\mathbf{C}} = &\mathbf{C} + \operatornamewithlimits{softmax}_{HW}(\mathbf{Q}^{c} \times (\mathbf{K}^{p})^{\mathrm{T}}) \times \mathbf{V}^{p} + \mathbf{Z}^{\mathrm{T}} \times \mathbf{V}^{p} \\
= & \mathbf{C} + (\operatornamewithlimits{softmax}_{HW}(\mathbf{Q}^{c} \times (\mathbf{K}^{p})^{\mathrm{T}}) + \mathbf{Z}^{\mathrm{T}}) \times \mathbf{V}^{p},
\end{split}
\end{equation}
where $\mathbf{Z}$ is obtained from \equref{eq:residual_cluster_transformer_perform_clustering}. The update is shown in the center panel of \figref{fig:detailed_clusteringattn}, while the effect of {\it densified} attention could be found in ~\figref{fig:clusterview_of_seg_new}.

\vspace{0.5ex}
\noindent\textbf{Solution to Infrequent Pixel Updates.} We propose to also utilize the clustering result $\mathbf{Z}$ to perform an update on the pixel features using the features of cluster centers, \ie,
\begin{align}
\label{eq:cluster_update_pixel}
\hat{\mathbf{F}} = &\mathbf{F} + \mathbf{Z} \times \mathbf{V}^{c},
\end{align}
where $\mathbf{V}^c \in \mathbb{R}^{N \times D}$ is the linearly projected values from the cluster centers. This update is performed within each stacked transformer decoder, enabling frequent communication between pixel features and cluster centers (right panel of \figref{fig:detailed_clusteringattn}).

To this end, we have improved the transformer cross-attention module by simultaneously updating the clustering result (\ie, pixel-cluster assignment), pixel features, and cluster centers. However, we notice that during the interaction between pixel features and cluster centers, pixel features are treated as bag of words~\cite{lazebnik2006beyond}, while the location information is not well utilized. Although learnable positional encodings (\ie object queries~\cite{carion2020end}) are used for the cluster center embeddings, the positional encodings are fixed for all input images, which is suboptimal when an object query predicts masks at different locations in different input images. To resolve the issue, we propose to adopt a dynamic positional encoding conditioned on the inputs for {\it location-sensitive} clustering.

\vspace{0.5ex}
\noindent\textbf{Location-Sensitive Clustering.} To inject dynamic location information to cluster centers, we explicitly predict a reference mask that consists of $M$ points for each cluster center. In particular, a MLP is used to predict the reference mask out of cluster center features, followed by a sigmoid activation function. That is, we have:
\begin{align}
\label{eq:ref_mask_pred}
&\hat{e} = e + \operatorname{MLP}(\mathbf{C}),\\
&r^{c} = \operatorname{sigmoid}(\hat{e}),
\end{align}
where $e \in \mathbb{R}^{N\times 2M}$ denotes an embedding projected from the cluster centers, and $r^{c}=[r^{c,h}, r^{c,w}]  \in \mathbb{R}^{N\times 2M}$ are the reference mask represented with $M$ pairs of coordinates $(r_i^{c,h}, r_i^{c,w})$. We utilize a residual update manner~\cite{he2016deep,zhu2020deformable} to predict the reference mask, with a skip-connection on the projected embedding $e$ across stages. The location space is normalized to $[0,1]\times[0,1]$.

We add location information to pixel features and cluster centers through a coordinate convolution~\cite{liu2018intriguing}. Specifically, we apply coordinate convolutions at the beginning of each transformer layer to ensure location information is considered during the clustering process, as shown below.
\begin{align}
\label{eq:red_mask_coordconv}
\hat{\mathbf{C}} &= \operatorname{Conv}(\operatorname{Concat}(\mathbf{C}, r^{c})),\\
\hat{\mathbf{F}} &= \operatorname{Conv}(\operatorname{Concat}(\mathbf{F}, r^{p})),
\end{align}
where $r^{p} \in \mathbb{R}^{HW\times 2}$ is the coordinates normalized to $[0,1]$ for pixels in image space, which is fixed and not learnable.

We note that compared to the reference point used in the Deformable DETR~\cite{zhu2020deformable}, the proposed reference mask provides a rough mask shape prior for the whole object mask. Besides, we adopt a much simpler way to incorporate the location information via coordinate convolution.

In order to learn meaningful reference mask predictions, we optimize the reference masks towards ground truth masks by proposing a mask approximation loss.

\vspace{0.5ex}
\noindent\textbf{Mask Approximation Loss.} We propose a loss to minimize the distance between the distribution of predicted reference points and that of points of ground-truth object masks. In detail, we utilize the Hungarian matching result to assign the ground-truth mask for each cluster center. Given the predicted $M$ points for each cluster center, we infer their extreme points~\cite{papadopoulos2017extreme} and mask center. We then apply an $L_1$ loss to push them to be closer to their ground-truth extreme points and center. Specifically, we have

{\footnotesize
\begin{align}
\label{eq:location_loss}
\notag &\mathcal{L}_{\mathrm{ext}} = \frac{1}{4K}\sum_{i=1}^{K}(|\operatorname{min}(r^{c,h}_{i}) - \operatorname{min}(y^{h}_{i})| + |\operatorname{max}(r^{c,h}_{i}) - \operatorname{max}(y^{h}_{i})|\\
\notag&~~~~~~~~~~~~~~~~~~~~~~+|\operatorname{min}(r^{c,w}_{i}) - \operatorname{min}(y^{w}_{i})| + |\operatorname{max}(r^{c,w}_{i}) - \operatorname{max}(y^{w}_{i})|),\\
&\notag\mathcal{L}_{\mathrm{cen}} = \frac{1}{2K}\sum_{i=1}^{K}(|\operatorname{avg}(r^{c,h}_{i}) - \operatorname{avg}(y^{h}_{i})| + |\operatorname{avg}(r^{c,w}_{i}) - \operatorname{avg}(y^{w}_{i})|),\\
&\mathcal{L}_{\mathrm{loc}} = \mathcal{L}_{\mathrm{ext}} + \mathcal{L}_{\mathrm{cen}},
\end{align}
}%
where $y=[y^h, y^w]$ are pixels on ground-truth masks and predicted reference masks have been filtered and re-ordered based on Hungarian matching results.

Finally, combining all the proposed designs results in our Clustering Mask Transformer, or CMT-DeepLab, which rethinks the current mask transformer design from the clustering perspective.

\subsection{Network Instantiation}
We instantiate CMT-DeepLab on top of MaX-DeepLab-S~\cite{wang2021max} (abbreviated as MaX-S). We first refine its architecture design. Afterwards, we enhance it with the proposed Clustering Mask Transformers.

\vspace{0.5ex}
\noindent\textbf{Base Architecture.} We use MaX-S~\cite{wang2021max} as our base architecture. To better align it with other state-of-the-art architecture designs~\cite{liu2021swin}, we use GeLU~\cite{hendrycks2016gaussian} activation to replace the original ReLU activation functions. Besides, we remove all transformer blocks in the pretrained backbones, which reverts the backbone from MaX-S back to Axial-ResNet-50~\cite{wang2020axial}.
On top of the backbone, we append six dual-path axial-transformer blocks~\cite{wang2021max} (three at stage-5 w/ channels 2048, and the other three at stage-4 w/ channels 1024), yielding totally six axial self-attention and six cross-attention modules, which aligns with the number of attention operations used in other works~\cite{carion2020end,cheng2021per}.
Additionally, we obtain a larger network backbone by scaling up the number of blocks in stage-4 of the backbone~\cite{swidernet_2020}. As a result, two different model variants are used: one built upon Axial-ResNet-50 backbone with number of blocks $[3,4,6,3]$ (starting from stage-2), and another built upon Axial-ResNet-104 with number of blocks $[3,4,24,3]$. See the supplementary material for a detailed illustration.

\vspace{0.5ex}
\noindent\textbf{Loss Functions.} Following~\cite{wang2021max}, we use the PQ-style loss and three other auxiliary losses for the model training, including the instance discrimination loss, mask-ID cross-entropy, and semantic segmentation loss. However, we note that the instance discrimination loss proposed in~\cite{wang2021max} aims to push pixel features to be close to the feature center computed based on the ground-truth mask,
instead of directly to the cluster centers.
Therefore, we adopt the pixel-wise instance discrimination loss, which learns closely aligned representations for all pixels from the same class, allowing better clustering results.

Formally, we sample a set of pixels $A$ from the image, where we add bias to pixels' sampling probability based on the size of object mask they belong to. Thus, final sampled pixels are more balanced from objects with different scales. Afterwards, we directly perform contrastive loss on top of these pixels with multiple positive targets~\cite{khosla2020supervised}:
\begin{align}
\label{eq:pixel_insdis_loss}
\mathcal{L}^{insdis}=\sum_{a\in A}\frac{-1}{|P(a)|}\sum_{p\in P(a)} \log \frac{\exp{\left(f_{a}\cdot f_{p} / \tau\right)}}{\sum_{b\in A} \exp{\left(f_{a}\cdot f_{b} / \tau\right)}},
\end{align}
where $P(a)$ is a subset of pixels of $A$ that belongs to the same cluster (\ie, object mask) with $a$, and $|P(a)|$ is its cardinally. We use $f$ to denote a pixel feature vector, and $\tau$ is the temperature.

\vspace{0.5ex}
\noindent\textbf{Recursive Feature Network.}
Motivated by DetectoRS~\cite{qiao2020detectors} and CBNet~\cite{liu2020cbnet}, we adopt a simple strategy, named Recursive Feature Network (RFN), to increase the network capacity by stacking twice the whole model (including the backbone and added transformer blocks). There are two main differences. First, since we do not employ an FPN~\cite{lin2017feature} (as in~\cite{qiao2020detectors}), we simply connect the features at stride 4 (\ie, same stride as the segmentation output). Second, we do not use the complicated fusion module proposed in~\cite{qiao2020detectors}, but simply average the features between two stacked networks, which we empirically found to be better by around 0.2\% PQ.

\label{sec:methods}
\begin{table*}[!t]
\centering
\small
\tablestyle{9pt}{1.1}
\begin{tabular}{l|c|c|c|ccc|ccc}
 &  &  &  & \multicolumn{3}{c}{val-set} & \multicolumn{3}{c}{test-dev}\\
method & backbone & TTA & params & PQ & PQ\textsuperscript{Th} & PQ\textsuperscript{St} & PQ & PQ\textsuperscript{Th} & PQ\textsuperscript{St}\\
 \shline
\multicolumn{10}{c}{box-based panoptic segmentation methods}\\
\hline
Panoptic-FPN~\cite{kirillov2019panoptic} & R101 & & & 40.3 & 47.5 & 29.5 & - & - & - \\
UPSNet~\cite{xiong2019upsnet} & R50 & & & 42.5 & 48.5 & 33.4 & - & - & - \\
UPSNet~\cite{xiong2019upsnet} & R50 & \checkmark & & 43.2 & 49.1 & 34.1 & - & - & - \\
UPSNet~\cite{xiong2019upsnet} & DCN-101~\cite{dai2017deformable} & \checkmark & & - & - & - &  46.6 & 53.2 & 36.7 \\
DETR~\cite{carion2020end} & R101 & & 61.8M & 45.1 & 50.5 & 37.0 & 46.0 & - & -\\
DetectoRS~\cite{qiao2020detectors} & RX-101~\cite{xie2017aggregated} & \checkmark & & - & - & - & 49.6 & 57.8 & 37.1 \\
\hline
\multicolumn{10}{c}{center-based panoptic segmentation methods}\\
\hline
Panoptic-DeepLab~\cite{cheng2019panoptic} & X-71~\cite{chollet2016xception} & & 46.7M & 39.7 & 43.9 & 33.2 & - & - & - \\
Panoptic-DeepLab~\cite{cheng2019panoptic} & X-71~\cite{chollet2016xception} & \checkmark & 46.7M & 41.2 & 44.9 & 35.7 & 41.4 & 45.1 & 35.9 \\
Axial-DeepLab-L~\cite{wang2020axial} & AX-L~\cite{wang2020axial} & & 44.9M & 43.4 & 48.5 & 35.6 & 43.6 & 48.9 & 35.6 \\
Axial-DeepLab-L~\cite{wang2020axial} & AX-L~\cite{wang2020axial} & \checkmark & 44.9M & 43.9 & 48.6 & 36.8 & 44.2 & 49.2 & 36.8 \\
\hline
\multicolumn{10}{c}{end-to-end panoptic segmentation methods}\\
\hline
MaX-DeepLab-S~\cite{wang2021max} & MaX-S~\cite{wang2021max} & & 61.9M & 48.4 & 53.0 & 41.5 & 49.0 & 54.0 & 41.6 \\
MaX-DeepLab-L~\cite{wang2021max} & MaX-L~\cite{wang2021max} & & 451M & 51.1 & 57.0 & 42.2 & 51.3 & 57.2 & 42.4 \\
MaskFormer~\cite{cheng2021per} & Swin-B$^\ddagger$~\cite{liu2021swin} & & 102M & 51.8 & 56.9 & 44.1 & - & - & - \\
MaskFormer~\cite{cheng2021per} & Swin-L$^\ddagger$~\cite{liu2021swin} & & 212M & 52.7 & 58.5 & 44.0 & 53.3 & 59.1 & 44.5 \\
\hline \hline
CMT-DeepLab & Axial-R50$^\ddagger$~\cite{wang2020axial} & & 94.9M & 53.0 & 57.7 & 45.9 & 53.4 & 58.3 & 46.0 \\
CMT-DeepLab & Axial-R104$^\ddagger$ & & 135.2M & 54.1 & 58.8 & \textbf{47.1} & 54.5 & 59.6 & 46.9 \\
CMT-DeepLab & Axial-R104$^\ddagger$-RFN & & 270.3M & 55.1 & 60.6 & 46.8 & 55.4 & 61.0 & \textbf{47.0} \\
CMT-DeepLab (iter 200k) & Axial-R104$^\ddagger$-RFN & & 270.3M & \textbf{55.3} & \textbf{61.0} & 46.6 & \textbf{55.7} & \textbf{61.6} & 46.8 \\
\end{tabular}
\vspace{-.8em}
\caption{Results comparison on COCO val and test-dev set. {\bf TTA:} Test-time augmentation. $\ddagger$: ImageNet-22K pretraining. We provide more comparisons with concurrent works in the supplementary materials.
}
\label{tab:coco_val_test}
\end{table*}

\begin{table*}[th]
\centering
\subfloat[
\textbf{CMT-DeepLab: clustering update}.
\label{tab:cmt_appearnace}
]{
\centering
\begin{minipage}{0.45\linewidth}{\begin{center}
\tablestyle{6pt}{1.1}
\begin{tabular}{l|ccc}
 & PQ & PQ\textsuperscript{Th} & PQ\textsuperscript{St} \\
\shline
\baseline{baseline} & \baseline{46.2} & \baseline{50.0} & \baseline{40.5}\\
+ clustering transformer & 47.1 & 51.0 & 41.1 \\
+ pixel-wise contrastive loss &\textbf{47.5} & \textbf{51.1} & \textbf{42.1} \\
\end{tabular}
\end{center}}\end{minipage}
}
\subfloat[
\textbf{CMT-DeepLab: location-senseitive clustering}.
\label{tab:cmt_location}
]{
\begin{minipage}{0.45\linewidth}{\begin{center}
\tablestyle{6pt}{1.1}
\begin{tabular}{l|ccc}
 & PQ & PQ\textsuperscript{Th} & PQ\textsuperscript{St} \\
\shline
\baseline{baseline} & \baseline{46.2} & \baseline{50.0} & \baseline{40.5}\\
+ ref. mask pred. & 46.6 & 50.3 & 40.9 \\
+ coord-conv & \textbf{46.9} & \textbf{50.6} & \textbf{41.3}\\
\end{tabular}
\end{center}}\end{minipage}
}
\\
\centering
\vspace{.3em}
\subfloat[
\textbf{CMT-DeepLab: architecture}.
\label{tab:cmt_arch}
]{
\begin{minipage}{0.5\linewidth}{\begin{center}
\tablestyle{4pt}{1.05}
\begin{tabular}{ccc|c|ccc}
clustering update & location & decoder & params &PQ & PQ\textsuperscript{Th} & PQ\textsuperscript{St} \\
\shline
\baseline{} & \baseline{} & \baseline{} & \baseline{61.9M} & \baseline{46.2} & \baseline{50.0} & \baseline{40.5}\\
\checkmark &  & & 61.9M & 47.5 & 51.1 & 42.1 \\
& \checkmark &  & 65.5M & 46.9 & 50.6 & 41.3 \\
 &  & \checkmark & 91.0M & 47.1 & 51.3 & 40.9\\
\checkmark &  & \checkmark & 91.0M & 48.1 & 51.9 & 42.2 \\
\checkmark & \checkmark & \checkmark & 94.9M & \textbf{48.4} & \textbf{52.1} & \textbf{42.8} \\
\end{tabular}
\end{center}}\end{minipage}
}
\subfloat[
\textbf{CMT-DeepLab: pretraining, post-processing, scaling}.
\label{tab:cmt_arch2}
]{
\begin{minipage}{0.45\linewidth}{\begin{center}
\tablestyle{4pt}{1.05}
\begin{tabular}{ccc|ccc}
ImageNet-22K & RFN & mask-wise merge  & PQ & PQ\textsuperscript{Th} & PQ\textsuperscript{St} \\
\shline
\baseline{} & \baseline{} & \baseline{} & \baseline{48.4} & \baseline{52.1} & \baseline{42.8}\\
\checkmark & & & 49.3 & 53.3 & 43.4\\
\checkmark & \checkmark & & 50.1 & \textbf{54.8} & 43.0 \\
\checkmark & \checkmark & \checkmark & \textbf{50.6} & \textbf{54.8} &\textbf{44.3}  \\
\end{tabular}
\end{center}}\end{minipage}
}
\vspace{-.1em}
\caption{CMT-DeepLab ablation experiments. Baseline is labeled with grey color. Results are reported in accumulative manner.}
\label{tab:ablations} \vspace{-.5em}
\end{table*}
\begin{table}[t]
\centering
\small
\tablestyle{8pt}{1.1}
\begin{tabular}{ccc|ccc}
res. & backbone & iters & PQ & PQ\textsuperscript{Th} & PQ\textsuperscript{St} \\
\shline
\baseline{641} & \baseline{Axial-R50} & \baseline{100k} & \baseline{50.1} & \baseline{53.5} & \baseline{44.9} \\
641 & Axial-R50 & 200k & 50.6 & 54.5 & 44.8 \\
\hline
1281 & Axial-R50 & 100k & 53.0 & 57.7 & 45.9 \\
1281 & Axial-R50 & 200k & 53.5 & 58.5 & 45.9 \\
\hline
641 & Axial-R104 & 100k & 51.7 & 55.4 & 46.4 \\
641 & Axial-R104 & 200k & 52.2 & 56.4 & 46.0 \\
1281 & Axial-R104 & 100k & 54.1 & 58.8 & 47.1 \\
1281 & Axial-R104-RFN & 100k & 55.1 & 60.6 & 46.8 \\
\end{tabular}
\vspace{-.8em}
\caption{Ablation on \textbf{input resolution/backbone/training iterations}. ImageNet-22K, mask-wise merge are used for all results.}
\label{tab:other_ablation}
\end{table}

\begin{figure*}[th]
    \centering
    \includegraphics[width=0.95\linewidth]{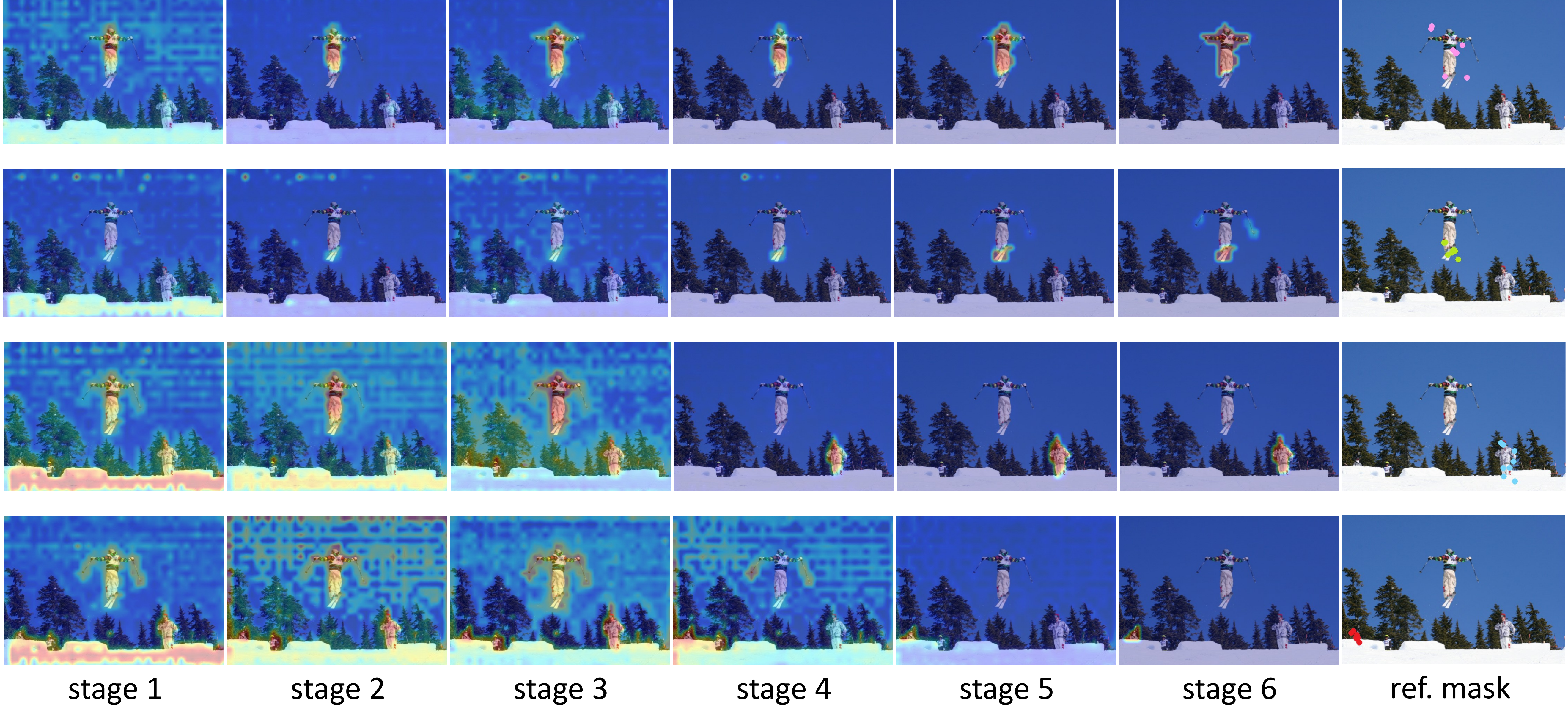}
    \caption{Visualization of clustering results at different stages (\ie, transformer layers), with last column for reference masks. The clustering results, providing denser attention maps, are close-to-random at the beginning and are gradually refined to focus on corresponding object.}
    \label{fig:clustering_vis}
\end{figure*}

\section{Experimental Results}

We report main results on COCO along with state-of-the-art methods, followed by ablation studies on the architecture variants, clustering mask transformers, pretrained weights, post-processing, and scaling strategies. Finally,
we analyze the working mechanism behind CMT-DeepLab with visualizations.

\vspace{0.5ex}
\noindent\textbf{Implementation Details.} We build CMT-DeepLab on top of MaX-DeepLab~\cite{wang2021max} with the official code-base~\cite{deeplab2_2021}. The training strategy mainly follows MaX-DeepLab. If not specified, the model is trained with 64 TPU cores for 100k iterations with the first 5k for warm-up. We use batch size = 64, Adam~\cite{kingma2014adam} optimizer, a poly schedule learning rate of $10^{-3}$. The ImageNet-pretrained~\cite{russakovsky2015imagenet} backbone has a learning rate multiplier 0.1. Weight decay is set to 0 and drop-path rate~\cite{huang2016deep} to 0.2. The input images are resized and padded to $1281\times 1281$ for training and inference. We use $|A|=4096$ for pixel-wise contrastive loss and $M=8$ for reference masks, we also tried other values but did not observe significant difference. Loss weight is 1.0 for the mask approximation loss. Other losses employ the same setting as~\cite{wang2021max}. During inference, we adopt a mask-wise merging scheme~\cite{cheng2021per} to obtain the final results.

\subsection{Main Results}
Our main results on the COCO panoptic segmentation \textit{val} set and \textit{test-dev set} are summarized in ~\tabref{tab:coco_val_test}.

\vspace{0.5ex}
\noindent\textbf{Val Set.} We compare our validation set results with box-based, center-based, and end-to-end panoptic segmentation methods. It is noticeable that CMT-DeepLab, built upon a smaller backbone Axial-ResNet-50, already surpasses all other box-based and center-based methods by a large margin. More importantly, when compared with its end-to-end baseline MaX-DeepLab-S~\cite{wang2021max}, we observe a significant improvement of 4.6\% PQ. Our small model even surpasses previous state-of-the-art method MaX-DeepLab-L~\cite{wang2021max}, which has more than $5\times$ parameters, by 1.9\% PQ. Compared to recently proposed MaskFormer~\cite{cheng2021per}, CMT-DeepLab still shows a significant advantage of 1.2\% PQ and 1.4\% PQ while being more light-weight over the small and large model variant, respectively. The significant improvement illustrates the importance of introducing the concept of clustering into transformer, which leads to a denser attention preferred by the segmentation task. Our CMT-DeepLab with a deeper backbone Axial-ResNet-104 improves the \textit{single-scale} performance to 54.1\% PQ, outperforming \textit{multi-scale} Axial-DeepLab~\cite{wang2020axial} by 10.2\% PQ. Moreover, we enhance the model with the proposed RFN, which further improves the PQ to 55.3\%.

\vspace{0.5ex}
\noindent\textbf{Test-dev Set.} We verify the transfer-ability of CMT-DeepLab on \textit{test-dev} set, which shows consistently better results compared to other methods. Especially, the small version of CMT-DeepLab with Axial-R50 backbone outperforms DETR~\cite{carion2020end} by 7.4\% PQ, MaX-DeepLab-S~\cite{wang2021max} by 4.4\% PQ, and MaX-DeepLab-L~\cite{wang2021max} by 2.1\% PQ. Additionally, employing a deeper backbone Axial-R104 can boost the PQ score by 1.1\% PQ. On top of it, using the proposed RFN further improves PQ to 55.7\%, surpassing MaskFormer~\cite{cheng2021per} with Swin-L~\cite{liu2021swin} backbone by 2.4\% PQ.

\subsection{Ablation Studies}
Herein, we evaluate the effectiveness of different components of the proposed CMT-DeepLab. For all the following experiments, we use MaX-DeepLab-S~\cite{wang2021max} with GeLU~\cite{hendrycks2016gaussian} activation function as our baseline. This improved baseline has a 0.3\% higher PQ compared to the original MaX-DeepLab-S. If not specified, we perform all ablation studies with the Axial-R50 backbone~\cite{he2016deep,wang2020axial}, ImageNet-1K~\cite{russakovsky2015imagenet} pretrained, crop size $641\times 641$, and $100$k training iterations.

\vspace{0.5ex}
\noindent\textbf{Clustering Mask Transformer.} We start with adding the design variants of Clustering Mask Transformer step by step, as summarized in Tab.~\ref{tab:cmt_appearnace}.
Regarding the object queries as cluster centers, and adding a clustering-style update can improve the PQ by 0.9\%, illustrating the effectiveness of the cluster center perspective and the importance of including more pixels into the cluster center updates. Next, we utilize pixel-wise contrastive loss instead of the original instance-wise contrastive loss, resulting in another 0.4\% PQ improvement, as it provides a better supervision signal from a clustering perspective. In short, re-designing the transformer layer from a clustering perspective leads to a 1.3\% PQ improvement overall.

\vspace{0.5ex}
\noindent\textbf{Location-Sensitive Clustering.} Location information plays an important role in the clustering process, as shown in Tab.~\ref{tab:cmt_location}. Each cluster center needs to predict a reference mask without using pixel features (\ie, appearance information), which requires cluster centers to include more location information in the feature embedding and thus benefits clustering. Adding reference masks prediction alone brings a gain of 0.4\% PQ.
Using the coordinate convolution (coord-conv)~\cite{liu2018intriguing} to include the reference mask information yields another 0.3\% PQ improvement.
In sum, the location-sensitive clustering brings up the PQ score by 0.7\%.

\vspace{0.5ex}
\noindent\textbf{Stronger Decoder.} We study the effect of using a stronger decoder design~\cite{carion2020end,cheng2021per}. We remove all transformer layers from the pretrained backbone, which reverts the MaX-S backbone~\cite{wang2021max} to Axial-ResNet-50~\cite{wang2020axial}. Then we stack more axial-blocks with transformer module in the decoder part. More specifically, we use six self-attention modules and six cross-attention modules in total for the decoder, which aligns to the design of DETR~\cite{carion2020end}. As shown in Tab.~\ref{tab:cmt_arch}, this stronger decoder brings 0.9\% PQ improvement (47.1\% $\vs$ 46.2\%).

As shown in Tab.~\ref{tab:cmt_arch}, these improvements are complementary to each other, while combining them together can further boost the performance. Adding all of them leads to CMT-DeepLab, which improves 2.2\% PQ over the MaX-DeepLab-S-GeLU baseline. We note that the major cost comes from the stronger decoder, which accounts for the increase of 29.1M parameters, while clustering update and location-sensitive clustering improve the PQ by 1.3\% and 0.7\%, respectively, with neglectable extra parameters.

\vspace{0.5ex}
\noindent\textbf{Pretraining, Post-processing, and Scaling.} We further verify the effect of better pretraining, post-processing, and scaling-up, with results summarized in Tab.~\ref{tab:cmt_arch2} and Tab.~\ref{tab:other_ablation}. Specifically, we find that using ImageNet-22K for pretraining can improve the performance by 0.9\% PQ. Furthermore, we empirically find that using the mask-wise merge strategy~\cite{cheng2021per} to obtain panoptic results, compared to the simple per-pixel strategy~\cite{wang2021max}, improves PQ by 0.5\%.
Next, we scale up CMT-DeepLab from different dimensions. With a longer training strategies (from 100k to 200k iterations), we observe a consistent 0.5\% PQ improvement over various settings, where the improvement mainly comes from PQ\textsuperscript{Th} (\ie, thing classes), indicating that the model needs a longer training schedule to better segment thing objects. We also find that using a larger input resolution (from 641 to 1281) significantly boosts the performance by more than 2\% PQ. Besides, increasing the model size by using a deeper backbone or stacking the model with RFN can improve the performance by 1.6\% and 1.0\%, respectively.

\vspace{0.5ex}
\noindent\textbf{Visualization.} In ~\figref{fig:clustering_vis}, we visualize  the clustering results in each stage as well as the learned reference masks.
As shown in the figure, the clustering results, starting with a close-to-random assignment, gradually learn to focus on the target instances. 
For example, in the last two rows of ~\figref{fig:clustering_vis}, the clustering results firstly focus on all the `person' instances and the background `snow', and then they start to concentrate on the specific person instance, showing a refinement from ``semantic segmentation" to ``instance segmentation".
Moreover, as shown in the last column of ~\figref{fig:clustering_vis}, the learned reference mask provides a reasonable prior for the object mask. 

\label{sec:experiments}
\vspace{-1ex}
\section{Conclusion}
\label{sec:conclusion}

In this work, we have introduced CMT-DeepLab, which rethinks object queries, used in the current mask transformers for panoptic segmentation, from a clustering perspective. Considering object queries as cluster centers, our framework additionally incorporates the proposed cluster center update in the cross-attention module, which significantly enriches the learned cross-attention maps and further facilitates the segmentation prediction. 
As a result, CMT-DeepLab achieves new state-of-the-art performance on the COCO dataset, and sheds light on the working mechanism behind mask transformers for segmentation tasks.

\paragraph{Acknowledgments.}
We thank Jun Xie for the valuable feedback on the draft. This work was supported in part by ONR N00014-21-1-2812.

\clearpage
{\small
\bibliographystyle{ieee_fullname}
\bibliography{egbib}
}
\clearpage

In the supplementary materials, we provide more technical details, along with more ablation and comparison results with other concurrent works. We also include more visualizations and comparisons over the baselines. Additionally, we provide a comprehensive comparison, in terms of training epochs, memory cost, parameters, FLOPs, and FPS, across different methods. We also report results with a ResNet-50 backbone for a fair comparison across different methods, along with additional results on Cityscapes. Finally, we summarize the limitations of our work and potential negative impacts.

\begin{figure*}[th]
    \centering
    \includegraphics[width=0.75\linewidth]{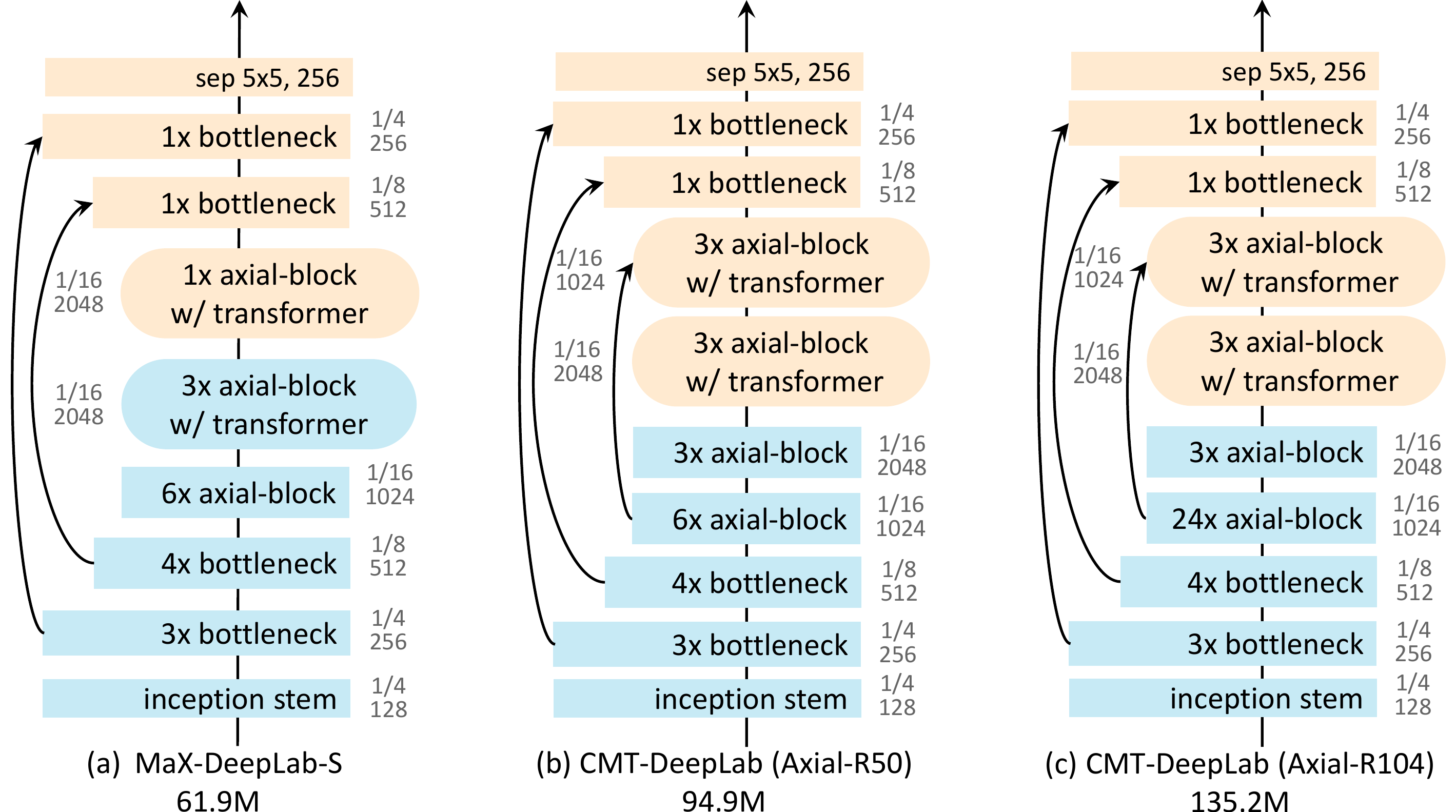}
    \caption{A visual comparison of architecture between MaX-DeepLab-S and CMT-DeepLab. Pretrained backbone part is labeled in blue color.
    }
    \label{fig:arch}
\end{figure*}

\section{More Technical Details}
\noindent\textbf{Backbones.} In \figref{fig:arch}, we provide an architectural comparison of MaX-DeepLab-S~\cite{wang2021max} and CMT-DeepLab built upon Axial-R50/104~\cite{wang2020axial}. Specifically, we simplify the backbone from MaX-DeepLab-S~\cite{wang2021max} by removing transformer modules in the backbone (light blue), and stacking more blocks in the decoder module (light orange). The Axial-R104 backbone is obtained by scaling up Axial-R50 (\ie, four times more layers in the stage-4).

\vspace{0.5ex}
\noindent\textbf{Recursive Feature Network.} We construct Recursive Feature Network (RFN) in a manner similar to~\cite{qiao2020detectors}. More specifically, we stack two models together, with a skip-connection from the decoder features at stride 4 in the first network to the encoder features at stride 4 in the second network. Instead of using the complicated fusion module proposed in~\cite{qiao2020detectors}, we simply average the features for fusion. Moreover, the two networks share the same set of cluster centers (\ie, object queries), which are sequentially updated from the first network to the second one. We also add supervision for the first network but use the Hungarian matching results based on the final output. 

\section{More Results and In-depth Analysis}
\noindent{\bf Effect of frequent pixel update (our second solution).}
As discussed in the main paper, the clustering results will be also used to update pixel features besides cluster centers to ensure a frequent pixel update. We tried removing the pixel feature updates from clustering transformer, which leads to a degradation of 0.4\% PQ.

\begin{table*}[!t]
\centering
\small
\tablestyle{7pt}{1.1}
\begin{tabular}{l|c|c|ccc|ccccc}
 &  &   & \multicolumn{3}{c}{val-set} & \multicolumn{5}{c}{test-dev}\\
method & backbone & params & PQ & PQ\textsuperscript{Th} & PQ\textsuperscript{St} & PQ & PQ\textsuperscript{Th} & PQ\textsuperscript{St} & SQ & RQ\\
 \shline
MaX-DeepLab-S~\cite{wang2021max} & MaX-S~\cite{wang2021max} & 61.9M & 48.4 & 53.0 & 41.5 & 49.0 & 54.0 & 41.6 & - & - \\
MaX-DeepLab-L~\cite{wang2021max} & MaX-L~\cite{wang2021max} & 451M & 51.1 & 57.0 & 42.2 & 51.3 & 57.2 & 42.4 & 82.5 & 61.3 \\
\hline
MaskFormer$^\dagger$~\cite{cheng2021per} & Swin-B$^\ddagger$~\cite{liu2021swin} & 102M & 51.8 & 56.9 & 44.1 & - & - & -  & - & - \\
MaskFormer$^\dagger$~\cite{cheng2021per} & Swin-L$^\ddagger$~\cite{liu2021swin} & 212M & 52.7 & 58.5 & 44.0 & 53.3 & 59.1 & 44.5 & 82.0 & 64.1 \\
K-Net$^\dagger$~\cite{zhang2021k} & R101-FPN~\cite{lin2017feature} & - &  49.6 & 55.1 & 41.4 & - & - & - & - & -  \\
K-Net$^\dagger$~\cite{zhang2021k} & R101-FPN-DCN~\cite{dai2017deformable} & - &  48.3 & 54.0 & 39.7 & - & - & - & - & -  \\
K-Net$^\dagger$~\cite{zhang2021k} & Swin-L$^\ddagger$~\cite{liu2021swin} & - &  54.6 & 60.2 & 46.0 & 55.2 & 61.2 & 46.2 & 82.4 & \textbf{66.1} \\
\hline \hline
CMT-DeepLab & Axial-R50$^\ddagger$~\cite{wang2020axial} & 94.9M & 53.0 & 57.7 & 45.9 & 53.4 & 58.3 & 46.0 & 83.0 & 63.6 \\
CMT-DeepLab & Axial-R104$^\ddagger$ & 135.2M & 54.1 & 58.8 & \textbf{47.1} & 54.5 & 59.6 & 46.9 & 83.2 & 64.7 \\
CMT-DeepLab & Axial-R104$^\ddagger$-RFN & 270.3M & 55.1 & 60.6 & 46.8 & 55.4 & 61.0 & \textbf{47.0} & 83.5 & 65.6 \\
CMT-DeepLab (iter 200k) & Axial-R104$^\ddagger$-RFN & 270.3M & \textbf{55.3} & \textbf{61.0} & 46.6 & \textbf{55.7} & \textbf{61.6} & 46.8 & \textbf{83.6} & 65.9 \\
\end{tabular}
\caption{Results comparison on COCO val and test-dev set. $\ddagger$: ImageNet-22K pretraining. $\dagger$: Concurrent works. We update comparison with concurrent works, and also our improved results with longer training iterations.
}
\label{tab:supp_coco_val_test}
\end{table*}

\noindent{\bf Comparison with more concurrent works.}
Also shown in ~\tabref{tab:supp_coco_val_test}, we compare our CMT-DeepLab with the baseline MaX-DeepLab~\cite{wang2021max}, and {\it concurrent} works MaskFormer~\cite{cheng2021per} and K-Net~\cite{zhang2021k} on the {\it test-dev} set.
As shown in the table, our best model (using 200K iterations and RFN) attains the performance of 55.7\% PQ on the test-dev set, which is 4.4\% and 2.4\% better than MaX-DeepLab-L~\cite{wang2021max} and MaskFormer~\cite{cheng2021per}. Our best model is 0.5\% PQ better than K-Net~\cite{zhang2021k}, which adopts a different framework (\ie, dynamic kernels) than mask-transformer-based approaches.
In addition to PQ, we further look into RQ and SQ for performance analysis. We observe that with a similar performance to K-Net~\cite{zhang2021k} in RQ, our best model performs better in SQ. Specifically, our best model yields 83.6\% SQ, which is 1.2\%, 1.6\%, and 1.1\% better than K-Net, MaskFormer, and MaX-DeepLab-L, respectively. Interestingly, our lightweight variant, CMT-DeepLab with Axial-R50, achieves 83.0\% SQ, which is still better than all the other methods. We attribute our better performance in SQ to the proposed clustering mask transformer layer, which yields denser attention maps to facilitate segmentation tasks.

\noindent{\bf Accuracy-cost Trade-off Comparison.}
We provide a comprehensive comparison of training cost (epochs, memory), model size (params, FLOPs, FPS), and performance (PQ) in~\tabref{tab:flops}.
The training memory is measured on a TPU-v4, while other statistics are measured with a Tesla V100-SXM2 GPU. We use TensorFlow 2.7, cuda 11.0, input size $1200\times 800$, and batch size $1$.
For MaskFormer (PyTorch-based), we cite the numbers from their paper.
As shown in the table, our CMT-DeepLab-S (Axial-R50) outperforms MaskFormer-SwinB by 1.2\% PQ with comparable model size and inference cost. Our CMT-DeepLab-S also outperforms MaskFormer-SwinL while using much fewer model parameters and running faster. All our models outperform MaX-DeepLab. Notably, our best model CMT-DeepLab-L-RFN (Axial-R104-RFN) outperforms MaX-DeepLab-L by 4.2\% PQ while using only 60\% model parameters and 33.6\% FLOPs.

\begin{table}[t]
\centering
\tablestyle{1.1pt}{1.1}
\begin{tabular}{l|cc|ccc|c}
method                & epochs & memory & params & FLOPs & FPS & PQ            \\ \shline
MaskFormer-SwinB~\cite{cheng2021per} & 300 & - & 102M  & 411G  & 8.4 & 51.8 \\
MaskFormer-SwinL~\cite{cheng2021per}  & 300 & - & 212M  & 792G  & 5.2  & 52.7 \\
\hline
MaX-DeepLab-S~\cite{wang2021max}    & 216   & 6.3G   & 62M  & 291G   & 11.9 & 48.4          \\
MaX-DeepLab-L~\cite{wang2021max}    & 216   & 28.7G  & 451M   & 3317G  & 2.2  & 51.1          \\
\hline
CMT-DeepLab-S    & 54   & 10.2G  & 95M  & 396G & 8.1  & 53.0          \\
CMT-DeepLab-L    & 54   & 11.8G  & 135M & 553G  & 6.0  & 54.1          \\
CMT-DeepLab-L-RFN & 54  & 25.8G  & 270M & 1114G & 3.2    & 55.1          \\
CMT-DeepLab-L-RFN & 108 & 25.8G   & 270M & 1114G & 3.2    & 55.3          \\
\end{tabular}
\caption{A comprehensive accuracy-cost trade-off comparison.
}
\label{tab:flops}
\end{table}

\noindent{\bf Backbone Differences.}
As different backbones are adopted for different methods (\eg, MaX-S/L~\cite{wang2021max}, Swin~\cite{liu2021swin}), it hinders a direct and fair comparison across different methods. To this end, we provide results based on a ResNet-50 backbone across different models on COCO \textit{val} set. As shown in~\tabref{tab:r50}, our CMT-DeepLab significantly outperforms MaX-DeepLab and concurrent works (MaskFormer and K-Net).

\begin{table}[t]
\centering
\tablestyle{2pt}{1}
\begin{tabular}{c|cccc}
 & MaskFormer~\cite{cheng2021per} & K-Net~\cite{zhang2021k} & MaX-DeepLab~\cite{wang2021max} & CMT-DeepLab \\ \shline
PQ     & 46.5       & 47.1  & 46.0           & \textbf{48.5}  \\
\end{tabular}
\caption{Results comparison with ResNet-50 as the backbone.}
\label{tab:r50}
\end{table}

\noindent{\bf Results on Cityscapes.}
We provide additional results on Cityscapes in~\tabref{tab:cityscapes}.
For a fair comparison, we adopt the \textbf{same} setting, including pretrain weights (IN-1k), training hyper-parameters (e.g., iterations 60k, learning rate 3e-4, crop size $1025\times 2049$), and post-processing scheme (pixel-wise argmax as in MaX-DeepLab). As shown in the table, our CMT-DeepLab-S significantly outperforms MaX-DeepLab-S by 2.9\% PQ and 1.6\% mIoU.

\begin{table}[t]
\centering
\tablestyle{4pt}{1.1}
\begin{tabular}{l|ccc|c}
method        & PQ & RQ & SQ & mIoU \\ \shline
\baseline{MaX-DeepLab-S} &   \baseline{61.7}  &  \baseline{74.5}  &  \baseline{81.5}  & \baseline{79.8}\\
CMT-DeepLab-S &   64.6  &  77.4  &  82.6  & 81.4 \\
\end{tabular}
\caption{Cityscapes \textit{val} set results.
}
\label{tab:cityscapes}
\end{table}

\section{Visual Comparison}

\noindent\textbf{Visualization Details.} To visually compare the clustering results/attention maps, we firstly follow DETR~\cite{carion2020end} to average values across multi-heads to obtain a single attention map, which is then transformed into a heatmap in a manner similar to CAM~\cite{zhou2016learning} by normalizing the values to the range [0, 255]. Note that we do not apply any smoothing techniques (\eg, square root), which in fact adjust the learned attention values. These differences make the visualization differ from those in the paper of MaX-DeepLab~\cite{wang2021max}. All visualizations are done with CMT-DeepLab based on Axial-R50, and MaX-DeepLab-S, with input size $641\times 641$.

\begin{figure*}[thb]
    \centering
    \includegraphics[width=1.0\linewidth]{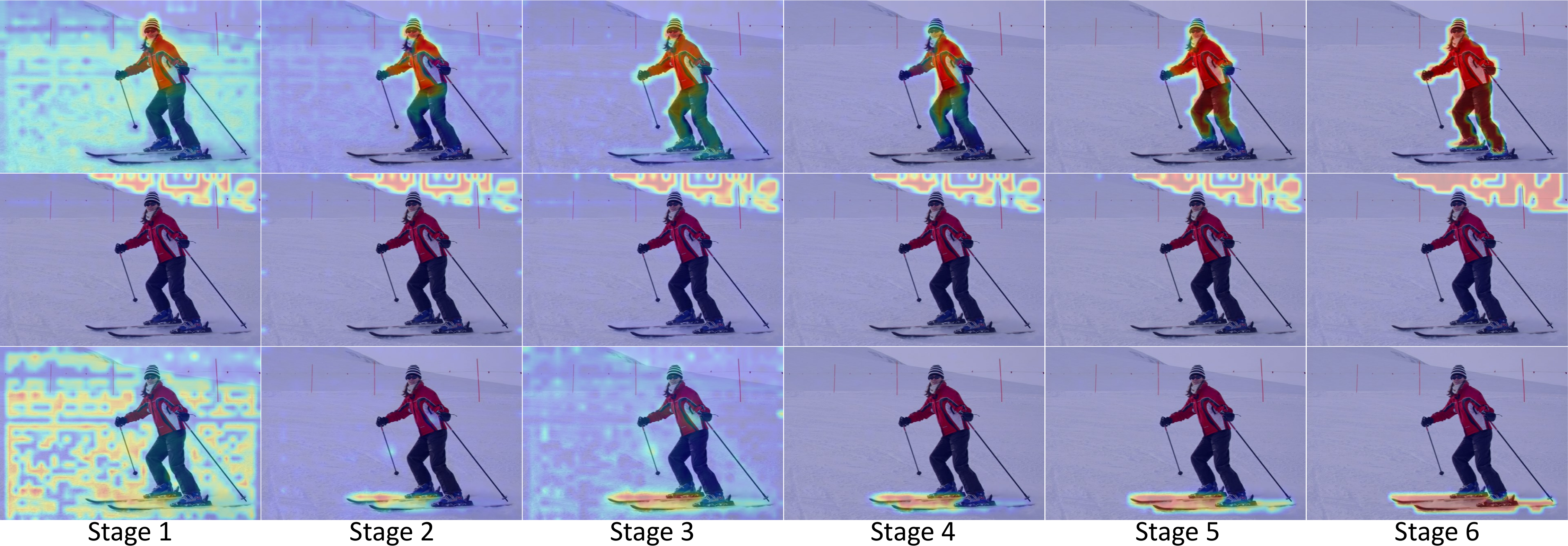}
    \caption{Visualization of clustering results at different stages (\ie, transformer layers). We note that clustering results for person (row 1) and skis (row 3) start from a close-to-random distribution at the beginning and are gradually refined to focus on corresponding target. But we also find some cluster centers, \eg, sky in row 2, are specialized in some semantic classes and start at a good semantic clustering.}
    \label{fig:supp_attn0}
\end{figure*}

\begin{figure*}[tbh]
    \centering
    \includegraphics[width=1.0\linewidth]{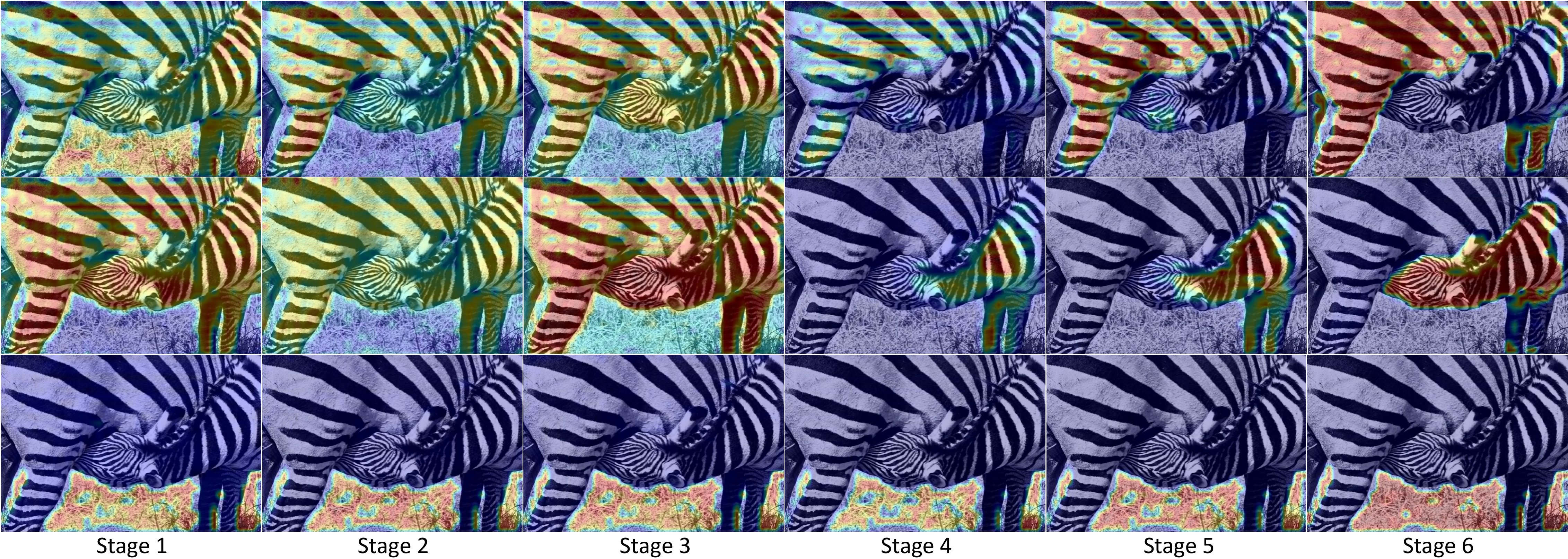}
    \caption{Visualization of clustering results at different stages (\ie, transformer layers). Both row 1 and 2 experience a semantic-to-instance refinement during the clustering process (\eg, in col 3, both clustering results capture all zebras.), which finally falls onto corresponding zebra. The cluster center on row 3 initializes with a good clustering result for grass, which coincides with the observation that some cluster centers intrinsically embed semantic information.}
    \label{fig:supp_attn2}
\end{figure*}

\begin{figure*}[tbh]
    \centering
    \includegraphics[width=1.0\linewidth]{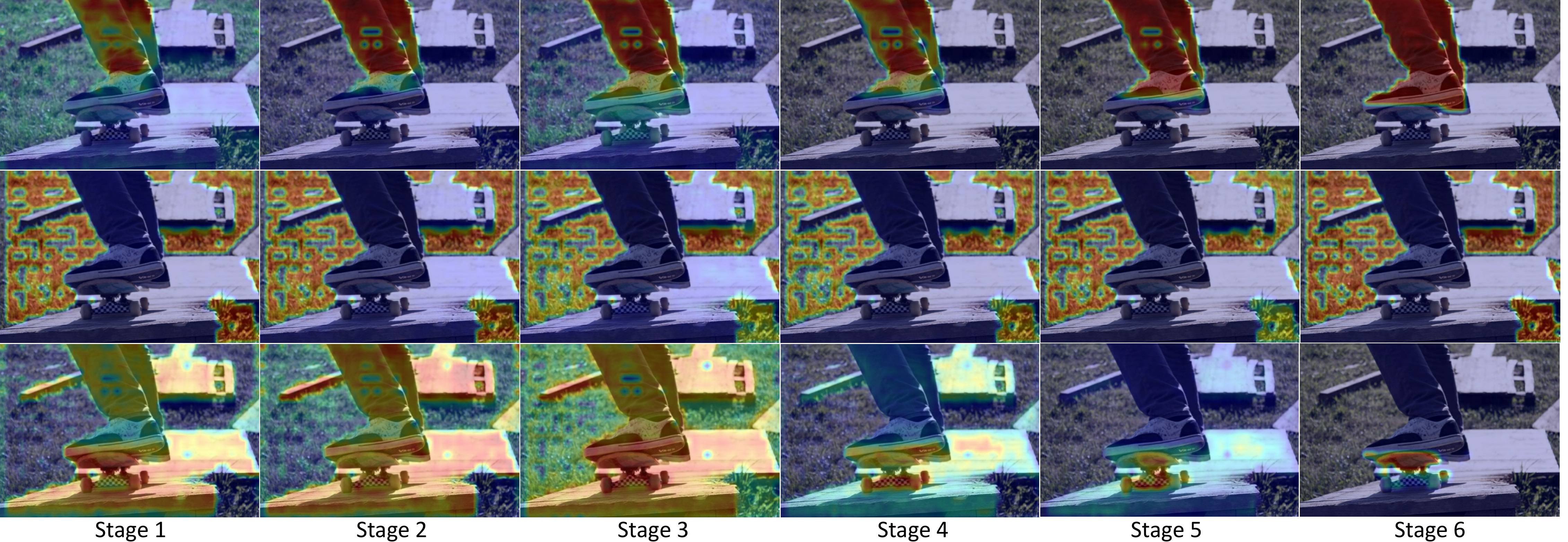}
    \caption{Visualization of clustering results at different stages (\ie, transformer layers). Row 1, 3 gradually falls into the target person and skateboard, while row 2 starts with a good clustering for grass.}
    \label{fig:supp_attn3}
\end{figure*}

\begin{figure*}[tbh]
    \centering
    \includegraphics[width=1.0\linewidth]{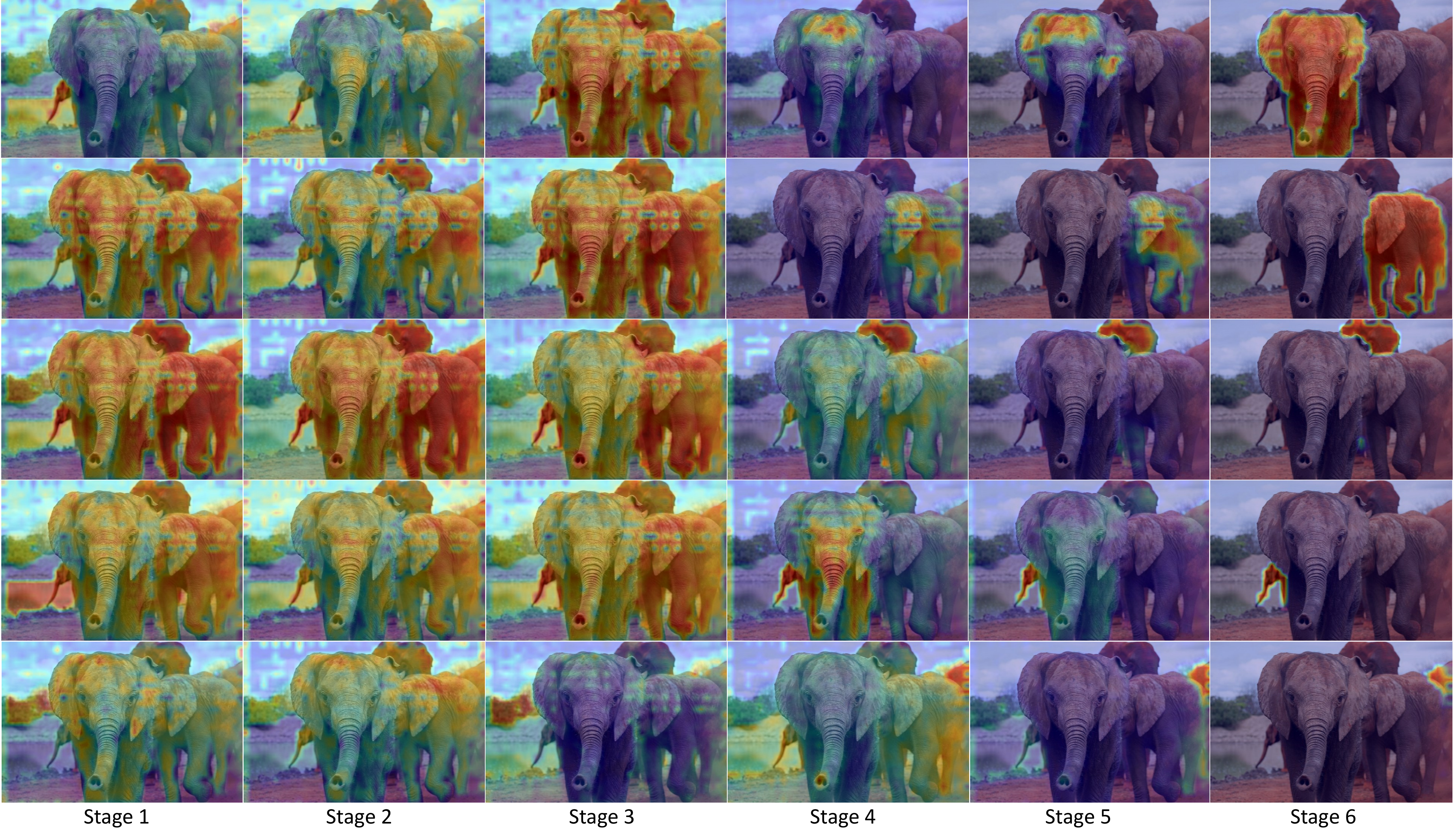}
    \caption{Visualization of clustering results at different stages (\ie, transformer layers). Each row corresponds to an elephant instance prediction. Similarly, most results start from a close-to-random clustering and gradually converge to the target in a semantic-to-instance manner.}
    \label{fig:supp_attn4}
\end{figure*}

\noindent\textbf{Clustering results.} In ~\figref{fig:supp_attn0},~\figref{fig:supp_attn2},~\figref{fig:supp_attn3}, and~\figref{fig:supp_attn4}, we provide more clustering visualization results. We observe the same trend as we presented in the main paper that the clustering results, providing denser attention maps, are close-to-random at the beginning and are gradually refined to focus on different objects. Interestingly, we also observe some exceptions (see ~\figref{fig:supp_attn0},~\figref{fig:supp_attn2},~\figref{fig:supp_attn3}), where the clustering results start with a good semantic-level clustering, indicating that some cluster centers can embed semantic information and thus specialize in some classes.

\begin{figure*}[tbh]
    \centering
    \includegraphics[width=1.0\linewidth]{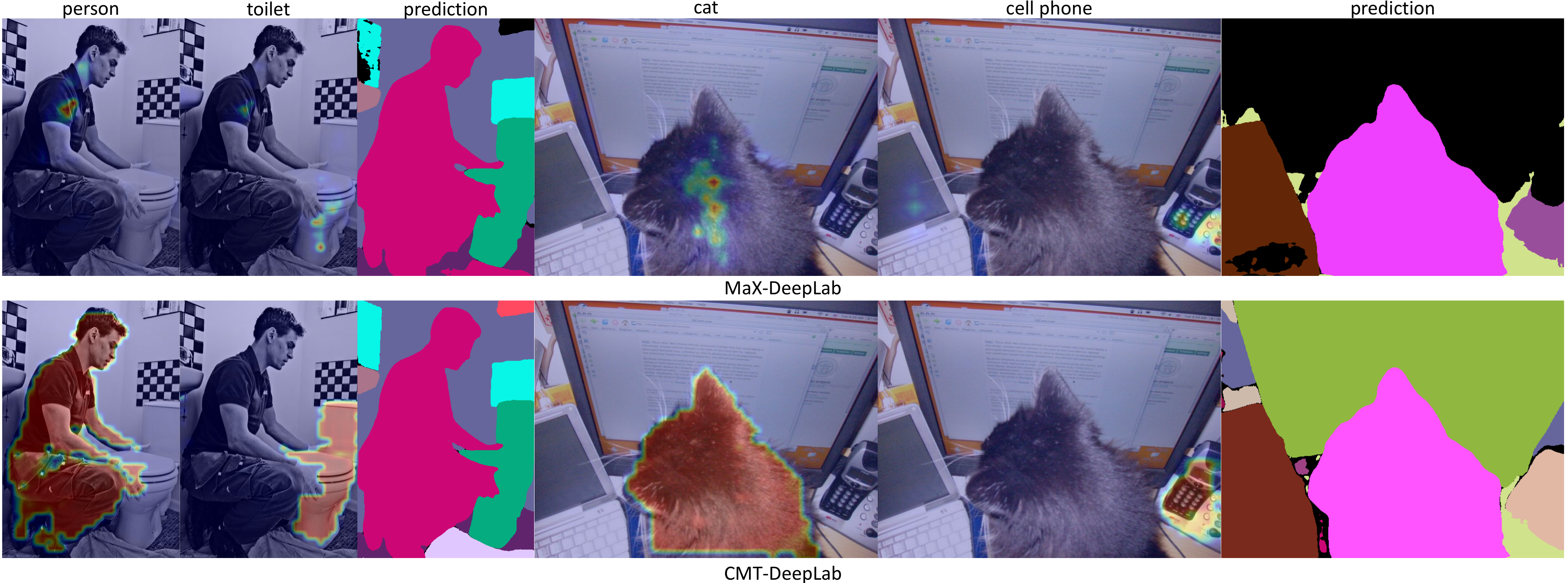}
    \caption{Visual comparison between CMT-DeepLab and MaX-DeepLab~\cite{wang2021max}. CMT-DeepLab provides a denser attention map to update cluster centers, which leads to superior performance in dense prediction tasks.
    }
    \label{fig:attn_comp0}
\end{figure*}

\begin{figure*}[tbh]
    \centering
    \includegraphics[width=1.0\linewidth]{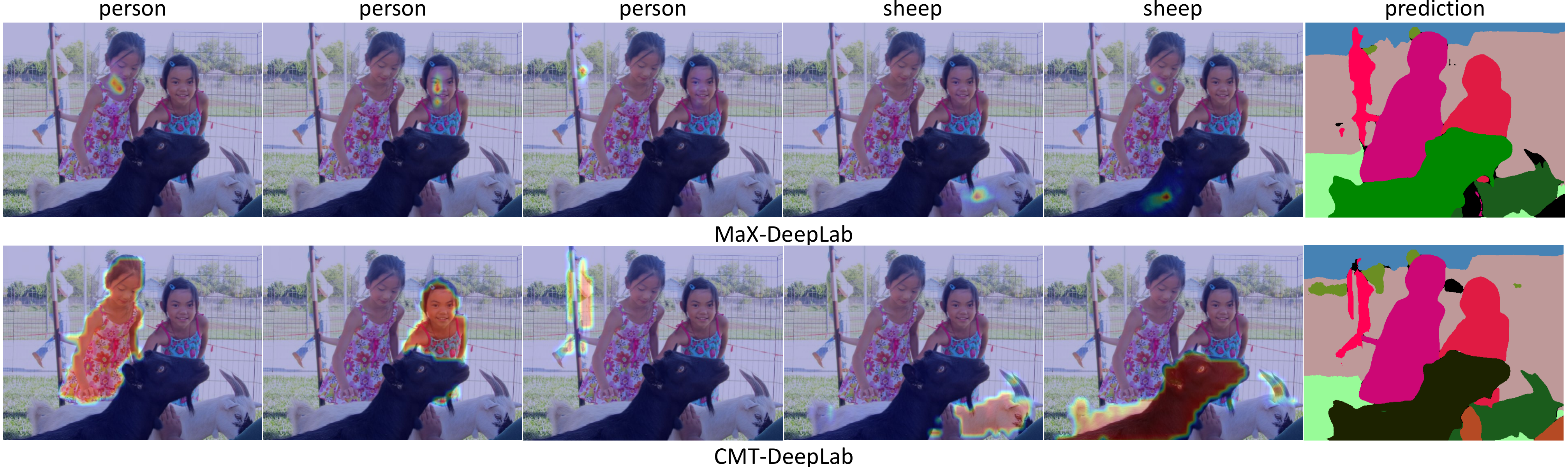}
    \caption{Visual comparison between CMT-DeepLab and MaX-DeepLab~\cite{wang2021max}. CMT-DeepLab provides a denser attention map to update cluster centers, which leads to superior performance in dense prediction tasks.
    }
    \label{fig:attn_comp1}
\end{figure*}

\begin{figure*}[tbh]
    \centering
    \includegraphics[width=1.0\linewidth]{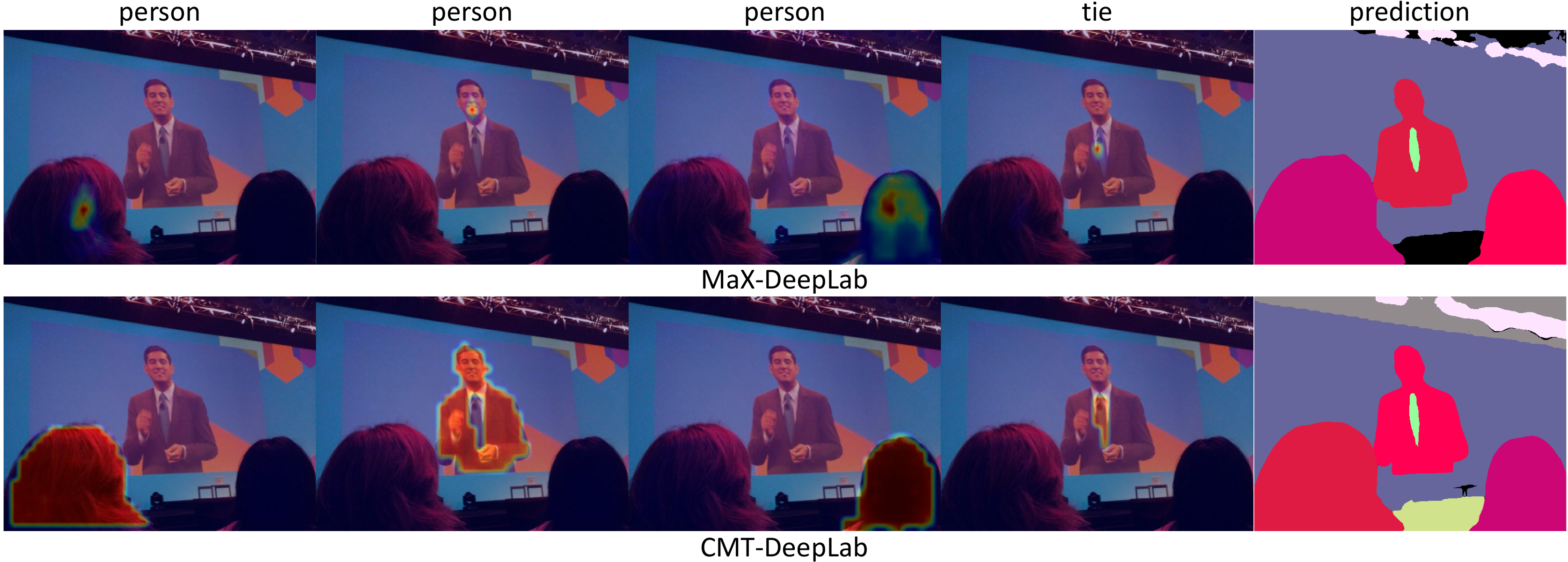}
    \caption{Visual comparison between CMT-DeepLab and MaX-DeepLab~\cite{wang2021max}. CMT-DeepLab provides a denser attention map to update cluster centers, which leads to superior performance in dense prediction tasks.
    }
    \label{fig:attn_comp2}
\end{figure*}

\noindent\textbf{Attention map comparison with MaX-DeepLab.}
In~\figref{fig:attn_comp0},~\figref{fig:attn_comp1}, and~\figref{fig:attn_comp2}, we show more attention map comparison with MaX-DeepLab. As shown in those figures, CMT-DeepLab provides a much denser attention map than MaX-DeepLab.

\section{Limitations}
Motivated from a clustering perspective, CMT-DeepLab generates denser attention maps and thus leads to a superior performance in the segmentation task.
However, the proposed clustering mask transformer, though significantly improves the segmentation quality (SQ), does not bring the same-level performance boost on the recognition ability (RQ).
Specifically, we have adopted some simple scaling-up strategies, including increasing model size, input size, or training iterations. Those strategies result in a large performance gain in RQ as a compensation, but with a cost at parameters, computation, or training time. It thus remains an interesting problem to explore in the future that how to improve its recognition ability efficiently and effectively.

\section{Potential Negative Impacts}
In this paper, we present a new panoptic segmentation framework, inspired by the traditional clustering-based algorithm, generates denser attention maps and further achieves new state-of-the-art performance. The findings described in this paper can potentially help advance the research in developing stronger, faster, and more elegant end-to-end segmentation methods. However, we also note that there is a long-lasting debate on the impacts of AI on human world. As a method improving the fundamental task in computer vision, our work also advances the development of AI, which means there could be both beneficial and harmful influences depending on the users.

\vspace{0.5ex}
\noindent\textbf{License of used assets.} COCO dataset~\cite{lin2014microsoft}: CC-by 4.0. ImageNet~\cite{russakovsky2015imagenet}: \url{https://image-net.org/download.php}.

\end{document}